\documentclass{article} 
\usepackage{iclr2026_conference,times}

\usepackage{amsmath,amsfonts,bm}

\def\eqref#1{equation~\ref{#1}}

\def\1{\bm{1}}

\DeclareMathAlphabet{\mathsfit}{\encodingdefault}{\sfdefault}{m}{sl}
\SetMathAlphabet{\mathsfit}{bold}{\encodingdefault}{\sfdefault}{bx}{n}

\usepackage{pdfpages}
\usepackage{hyperref}
\usepackage{url}
\usepackage{microtype}
\usepackage{graphicx}
\usepackage{subcaption}
\usepackage{booktabs}
\usepackage{multirow}
\usepackage{ulem}

\usepackage{amsmath}
\usepackage{amssymb}
\usepackage{mathtools}
\usepackage{amsthm}
\usepackage[capitalize,noabbrev]{cleveref}

\title{One Step Forward and K Steps Back: Better Reasoning with Denoising Recursion Models}

\usepackage{standalone}
\usepackage{tikz}
\usetikzlibrary{positioning,arrows.meta,shapes.geometric,calc,decorations.pathmorphing}
\usetikzlibrary{backgrounds, fit}
\tikzset{
    pics/noise_arrow/.style={
        code={
            \draw[thick, orange, line join=round, line cap=round, -Stealth] 
                (0,0) -- (0.2,0)              
                -- (0.35, 0.25) -- (0.5, -0.25) 
                -- (0.65, 0.25) -- (0.8, -0.25)
                -- (0.95, 0)                  
                -- (1.5, 0);                  
        }
    }
}

\usepackage{xcolor}
\usepackage{soul}
\usepackage{algorithm}
\usepackage{algpseudocode}
\usepackage{caption}
\usepackage{fontawesome5}

\definecolor{lightred}{RGB}{255,200,200}
\definecolor{lightgreen}{RGB}{200,255,200}
\definecolor{darkblue}{RGB}{0,0,139}
\definecolor{commentgray}{RGB}{128,128,128}

\newcommand{\hlred}[1]{{\sethlcolor{lightred}\hl{#1}}}
\newcommand{\hlgreen}[1]{{\sethlcolor{lightgreen}\hl{#1}}}
\newcommand{\kw}[1]{\textcolor{darkblue}{#1}}

\usepackage{listings}

\theoremstyle{plain}

\theoremstyle{definition}

\theoremstyle{remark}

\author{Chris Cameron$^{1}$, Wangzheng Wang$^{1~3}$, Nikita Ivanov$^{1~2}$, Ashmita Bhattacharyya$^{1~2}$,\\ \textbf{Didier Ch\'etelat$^{1}$ \& Yingxue Zhang$^{1}$} \\
$^{1}$Huawei Technologies, $^{2}$University of Toronto, $^{3}$University of Waterloo \\
\texttt{\{chris.cameron,wangzheng.wang2,ashmita.bhattacharyya,\}}\\
\texttt{\{didier.chetelat,yingxue.zhang\}}@huawei.com
}
\iclrfinalcopy 
\begin{document}

\maketitle
\begin{abstract}
Looped transformers scale computational depth without increasing parameter count by repeatedly applying a shared transformer block and can be used for iterative refinement, where each loop rewrites a full fixed-size prediction in parallel. On difficult problems, such as those that require search-like computation, reaching a highly structured solution starting from noise can require long refinement trajectories. Learning such trajectories is challenging when training specifies only the target solution and provides no supervision over the intermediate refinement path. 
Diffusion models tackle this issue by corrupting data with varying magnitudes of noise and training the model to reverse it in a \textit{single step}. However, this process misaligns training and testing behaviour. We introduce Denoising Recursion Models, a method that similarly corrupts data with noise but trains the model to reverse the corruption over \textit{multiple} recursive steps. This strategy provides a tractable curriculum of intermediate states, while better aligning training with testing and incentivizing non-greedy, forward-looking generation. Through extensive experiments, we show this approach outperforms the Tiny Recursion Model (TRM) on ARC-AGI, where it recently achieved breakthrough performance. \href{https://github.com/wwwwwwwwz/DenoisingRecursionModels}{\faGithub}
\end{abstract}

\section{Introduction}
Looped transformers, also known as recursive transformers, are a class of deep learning models that repeatedly apply the same transformer layer to an input to produce a prediction at test time \citep{giannou2023looped}. Tying weights across loops imposes an architectural constraint that encourages the model to learn update rules that generalize across depth, acting as an implicit regularizer that can improve generalization. Empirically, this constraint  is associated with stronger algorithmic reasoning \citep{dehghani2018universal, geiping2025scaling} as well as improved parameter efficiency \citep{saunshi2025reasoning} and sample efficiency \citep{zhu2025scalinglatentreasoninglooped}.

A novel looped transformer called the Tiny Recursion Model (TRM) \citep{jolicoeur2025less} recently made headlines by achieving breakthrough performance on ARC-AGI \citep{chollet2025arc}, a notoriously difficult general reasoning benchmark consisting of novel puzzles, each with transformation that must be inferred from only a handful of examples. This accomplishment was particularly dramatic as this model, composed of barely seven million parameters, outperformed leading approaches based on Large Language Models (LLMs) such as o3-high \citep{openaio3}, which has thousands of times more parameters and is pretrained at extreme computational cost.

The TRM is the direct successor of the Hierarchical Reasoning Model (HRM) \citep{wang2025hierarchical} and the latest installment in a line of \textit{iterative-refinement} looped transformers, dating back to \citet{lee2018deterministic}. Iterative-refinement models are non-autoregressive: at each loop they predict a complete, fixed-size output in parallel (one token/cell per position). Unlike chain-of-thought which appends newly generated tokens, these models are Markovian as subsequent loops rewrite the same-size prediction. They typically train by starting from noise and applying a sequence of recursions to match the target output. To borrow terminology from the \textit{diffusion} literature, we could call these models {\it backward-training} as they start from noise and try to undo it, as in the backward process of a diffusion model.

TRMs substantially outperform standard transformers on ARC-AGI-2, however they still leave the majority of ARC-AGI-2 problems unsolved. One explanation is that it is too demanding for the model to discover long, coherent trajectories starting from random noise and ending at highly-structured outputs without any additional supervision. This can be seen as analogous to sparse rewards in reinforcement learning (RL), where a random initial policy rarely stumbles onto a useful path to learn from. Long horizons becomes especially problematic because naively backpropagating through all recursive steps creates training instability \citep{he2016deep} and requires storing activations across the entire unrolled computation. In practice, many looped models \citep{geiping2025scaling,wang2025hierarchical}, including TRM, therefore rely on \textit{Truncated Backpropagation Through Time} (TBPTT) \citep{williams1995gradient}; they unroll the recursion for a fixed window of $k$ steps, backpropagate through that window, then detach the state and continue the unrolling. 

These long-horizon optimization issues motivate training objectives that avoid backpropagating through long unrolled recursions. Conditional diffusion models \citep[Chapter 8]{lai2025principles} provide a clean workaround: they sample intermediate states from a process of corrupting the prediction target and train a \textit{denoiser} model to reverse that corruption in a \textit{single step}. This provides an automatic easy-to-hard curriculum by varying the degree of noise, considerably simplifying the optimization landscape. Perhaps backward-training, without any form of curriculum learning, is simply too demanding on the more difficult problems. Diffusion is also an iterative-refinement model since the same denoiser is applied repeatedly at test time, albeit diffusion has largely developed separately from the looped-transformer literature. Since they start from the target output and add noise to it, we could call these models {\it forward-training} (in reference to the forward process in diffusion).

While forward-training offers a solution to long-horizon training, we report experimental results in Section \ref{section:results} showing that it substantially underperforms backward-training on ARC-AGI. We attribute this poor performance to a discrepancy between how the model is trained and how it is used at test time. During training, intermediate states are produced by noising the target and the model is optimized to remove that noise in a \textit{single step}. At test time, however, the model is iterated on its own predictions for many steps; the intermediate states it encounters can therefore differ substantially from the noised-target states seen in training, and the one-step objective does not incentivize learning to produce intermediate states that remain useful under further self-application. Training through a recursive window gives the model the capacity for \textit{implicit planning} \citep{guez2019investigation} in latent space before committing to a final answer. This train-test mismatch could lead to compounding errors and encourage myopic, greedy updates when the model is unrolled at inference. 

In contrast, backward-training methods avoid the train-test discrepancy. This points to a natural goal: incorporate the curriculum learning of forward-training while retaining the train-test alignment of backward-training. We therefore propose Denoising Recursion Models (DRMs), a novel looped transformer that takes one forward step along the diffusion corruption process into a corrupted target and then learns to recover the clean target over $k$ recursive denoising steps, rather than in a single jump. Crucially, our method only requires optimizing a single recursive window, rather than long episodes with a TBPTT detachment schedule. At test time, prediction proceeds as in TRM and diffusion by repeatedly applying the transformer from a maximally corrupted initialization. See Section~\ref{sec:methods} for full details.

In extensive experiments on ARC-AGI, using masking as the noise process, we found that DRMs outperformed TRMs at the same parameter count and surpassed the state of the art among open-source LLM-based baselines (NVARC, \citet{sorokin2025nvarc}) when controlling for training data. In addition, we investigate an alternative diffusion-TRM hybrid approach, which we call the State Perturbation Recursion Model (SPRM), where we inject noise into the intermediate latent states of the TRM procedure during training in order to explore more diverse trajectories. SPRM similarly improves upon TRM in small-data regimes, but it does not match DRM in larger-data regimes. In these data-rich settings, inherent data diversity likely makes the extra exploration from state perturbations unnecessary.

\section{Related Work}

Recurrent architectures are defined by the use of weight sharing to process sequences or generate computational depth. Historically, Recurrent Neural Networks (RNNs) \citep{elman1990finding} and LSTMs \citep{hochreiter1997long} applied recurrence along the sequence dimension, updating a fixed-size hidden state as they step through positions, which makes computation sequential and prevents parallelization across sequence positions. In contrast, looped transformers \citep{dehghani2018universal, geiping2025scaling} apply recurrence along the depth dimension: they process the entire sequence in parallel while iterating a shared layer to scale depth arbitrarily. In this section we provide additional details on this prior work.

\subsection{Looped Transformers}

\citet{dehghani2018universal} introduced the Universal Transformer, showing that recurrence across depth can improve reasoning while preserving parallelism across sequence positions. Subsequent work made this motivation more concrete in both theory and practice. \citet{selsam2018learning} showed that looped graph networks trained on SAT can generalize to harder instances simply by deeper unrolling. \citet{giannou2023looped} showed that looped transformers can emulate a small instruction-set computer, and \citet{merrill2025little} showed that these models can represent functions using only logarithmic depth in the number of steps required by discrete chain-of-thought. Empirically, looped models have shown improved parameter efficiency, sample efficiency, and length generalization on reasoning tasks \citep{saunshi2025reasoning,geiping2025scaling,fan2025loopedtransformerslengthgeneralization,zhu2025scalinglatentreasoninglooped}. These results make looped transformers a natural architecture for algorithmic reasoning.

Long-horizon recursive training creates a common memory and optimization problem: storing activations scales linearly with unrolled depth, while gradients degrade as the horizon grows. Existing responses include TBPTT for explicit finite-horizon training \citep{williams1995gradient}, one-step denoising objectives in diffusion, implicit differentiation in Deep Equilibrium Models (DEQ) \citep{bai2019deep,bai2021stabilizing}, and RL-style optimization over long reasoning trajectories in autoregressive and diffusion settings \citep{black2024diffusionrl,he2025mdpo}. We focus on a different point in this design space: preserving explicit multi-step differentiation through a short recursive window while avoiding the need to train from scratch over long trajectories.

Our work represents the latest in a line of research investigating \textit{small} iterative-refinement recursion models, which have demonstrated remarkably data-efficient reasoning without massive pretraining, particularly ARC-AGI. See Appendix \ref{app:background} for a detailed overview of ARC-AGI and the most successful solutions. The Hierarchical Reasoning Model (HRM) \citep{wang2025hierarchical} first demonstrated this potential by using alternating ``fast'' and ``slow'' recurrent steps, where the fast step is run many times for every execution of the slow step. Despite having only 27M parameters, HRM outperformed high-performance LLMs like o3-mini-high on ARC-AGI. This was simplified by the TRM \citep{jolicoeur2025less}, which uses a single homogeneous recursive layer. Cutting down to just 7M parameters, TRM outperformed HRM and surpassed 100B+ parameter models like o3-high. Very recently, the Universal Reasoning Model (URM) \citep{gao2025universal} extended TRM by applying a 1D convolution over the token sequences at the end of the MLP block, yielding an $\sim$8\% improvement on ARC-AGI-2. \citet{hu2025arc} (ViARC) used a similarly small but non-looped vision transformer---a 2$\times$2 patch encoding/decoding rather than encoding/decoding only at the token level---and improved performance over TRM. We think these TRM improvements of (1) changing encoding/decoding (ViARC) and (2) model architecture (URM) are fundamentally independent of our method; we anticipate that importing these ideas would see similar improvements.

\subsection{Diffusion}

Diffusion models are a type of looped, iterative-refinement model, first introduced by \citet{sohl2015deep}. They treat refinement as the reversal of a stochastic corruption process. 
\citet{ho2020denoising} introduced Denoising Diffusion Probabilistic Models (DDPM), which is the dominant formalism of diffusion today. While these foundations were established in continuous state-spaces, they can also be applied to tasks with discrete tokens. One approach bridges this gap by mapping discrete tokens into continuous embedding spaces---effectively latent diffusion---where standard Gaussian noise is applied and reversed before projecting back to discrete text \citep{li2022diffusion, gong2022diffuseq}.

Alternatively, diffusion can operate directly in discrete domains via \textit{stochastic transition matrices} \citep{austin2021structured} where tokens are corrupted by stochastically mapping tokens to a distribution over the vocabulary, or \textit{absorbing masking states} where the models learns to iteratively recover tokens from a masked input \citep{chang2022maskgit}. Recent methods like Llada \citep{nie2025llada,you2025llada,zhu2025llada} have shown that masked diffusion for training language models can be competitive with autoregressive models.

\subsection{Intersection of Diffusion and Recursion}

\citet{lee2018deterministic} introduced an early iterative-refinement model and trained it using a \textit{denoising-autoencoder} objective. In our terminology, their procedure stochastically interleaves forward-training (denoising a corrupted version of the target) and backward-training (refining from an uninformative initialization), sampling between the two modes with a mixture probability. Although this work predates diffusion terminology, the denoising-autoencoder component plays a similar role to diffusion by providing intermediate difficulty states via controlled corruption of the ground truth. Our DRMs are closely related but differ in how these ingredients are combined. Lee et al.'s forward-training updates supervise a single refinement step, whereas their backward-training updates still require traversing the trajectory from scratch. DRMs instead couple forward corruption with multi-step refinement within each example: we sample a corruption level, initialize the recursion at the corrupted target, and train to recover the clean output over a window of $k$ recursive steps.

Several recent papers explicitly frame looped models as diffusion-like processes. \citet{geiping2025efficient} 
build on \citet{chen2024diffusionforcingnexttokenprediction}'s work that proposed an asynchronous update schedule where the model processes token $k$ at loop depth $i-k$, effectively creating a \textit{diagonal frontier}. Unlike our methods, their primary focus is computational efficiency during inference and their experiments use pretrained looped models \citep{geiping2025scaling} that do not use diffusion objectives, whereas our approach is a training method. 
Similarly, \citet{zhou2025coevolutionarycontinuousdiscretediffusion} characterize looped models as continuous diffusion processes lacking intermediate noise and they propose a hybrid ``co-evolutionary'' architecture that combines discrete and continuous denoising.

An alternative to the backward-training methods used in standard iterative-refinement models is to optimize the denoising trajectory with reinforcement learning (RL) \citep{black2024diffusionrl}. \citet{he2025mdpo} recently applied this idea to masked diffusion, using intermediate rewards based on similarity to the ground truth. These rewards are analogous in spirit to the deep-supervision signals used in TRM, but the credit-assignment story is different: because the denoising policy is optimized over the full rollout, each update is valued according to how it contributes to eventual success under the same multi-step process used at inference. In this sense, RL directly aligns the training objective with test-time denoising, and \citet{he2025mdpo} report substantial improvements over LLaDA on mathematical and reasoning benchmarks.

A related idea appears in a different RL setting: recurrent policies acting in external MDPs rather than denoising trajectories. There, repeated internal computation has been discussed as a form of \textit{implicit planning}, where the policy can refine its latent state before committing to an action. \citet{guez2019investigation} introduced the Deep Repeated ConvLSTM (DRC), and \citet{bush2025} later provided evidence consistent with planning-like latent updates in this architecture. The analogy to DRMs is only partial, since we train with supervised losses rather than policy optimization. Still, backpropagating through multiple recursive denoising steps gives the model a related incentive to shape latent states for their downstream usefulness, rather than only for immediate denoising gain.

\section{Methods} \label{sec:methods}

\subsection{Preliminaries: Training Looped Models} In this work, we operate in the non-autoregressive sequence-to-sequence setting. We are given a dataset of training pairs $(X, Y) \sim \mathcal{D}$, where the input $X$ and ground truth $Y$ are sequences of length $M$. Both are represented as one-hot encodings over a vocabulary $\mathcal{V}$, such that $X, Y \in \{0,1\}^{M \times |\mathcal{V}|}$. Each unique token is mapped to a learned continuous embedding in $\mathbb{R}^{d}$. To generate predictions from a latent state $H\in \mathbb{R}^{M\times d}$, the model uses a decoder head to generate a probability distribution over the target tokens:
\begin{equation}
P_{\theta}(Y \mid H) = \text{softmax}(\text{dec}(H)).
\end{equation}
We focus on \textit{looped models}, a class of functions where the latent state is generated via a shared transition operator $\Phi_{\theta}$ applied recursively:
\begin{align}
H_{t} = \Phi_{\theta}(H_{t-1}, X, \dots), \quad t=1 \dots T
\end{align}
where $H_t$ is the latent state at recursion step $t$ and $H_0$ is an initialization distribution (typically Gaussian noise). $T$ represents the maximum recursive depth used during inference. While all looped models share this inference structure, they have different ways of training $\Phi_{\theta}$ to manage the training instability and memory constraints over a long horizon $T$.

Many of the methods considered here rely on a forward corruption process
$q_\tau(\tilde{Y}\mid Y)$, instantiated from a generic family
$\{q_\tau(\tilde{Y}\mid Y)\}_{\tau\in[0,1]}$, where $\tau$ denotes the
corruption level. This family may represent, for example, Gaussian diffusion
for continuous corruption \citep{ho2020denoising} or masking for discrete
corruption \citep{nichol2021improved}. We assume only that $\tau \sim \mathcal{U}(0,1)$, that $q_0(\tilde{Y}\mid Y)$ leaves the target
uncorrupted (i.e., $\tilde{Y}=Y$ with probability $1$), and that larger $\tau$ corresponds to progressively stronger corruption, with $\tau=1$
denoting maximal corruption.

\begin{figure*}[t]
    \centering
    \resizebox{\textwidth}{!}{\input{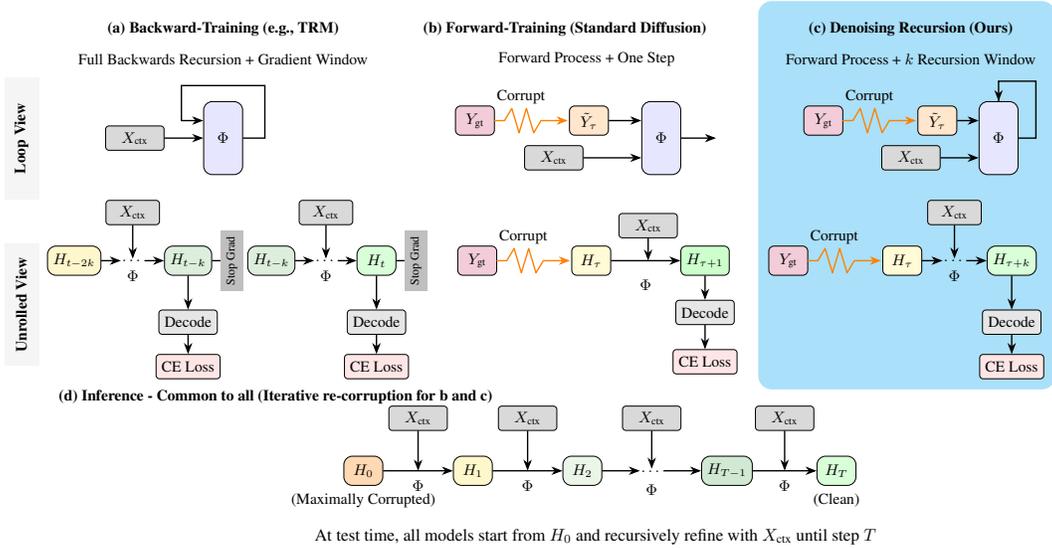}} 
    \caption{Architectural comparison between (a) backward-training (e.g., TRM), (b) forward-training (Standard Diffusion), and (c) DRMs. Like forward-training, DRMs sample the forward process and like backward-training, they train over $k$ recursions of transition operator $\Phi$. All methods recursively map from maximal corruption to a clean target during inference (d).}
    \label{fig:model_arch}
\end{figure*}

\subsubsection{Forward-Training Models (Diffusion)}
Diffusion models decouple the training procedure from the recursive generation process. A forward process $q_\tau(\tilde{Y}\mid Y)$ corrupts the ground truth $Y$ into a noisy state $\tilde{Y}_\tau$, effectively fast-forwarding the recursion process to intermediate states of varying levels of completeness. The model is trained to reverse this corruption in a \textit{single step}; $\Phi_{\theta}$ is applied exactly once to the noisy input $\tilde{Y}_\tau$, preventing gradients from flowing through time. Importantly, our objective is a reconstruction objective: the model is trained to predict the clean target $Y$ from $\tilde{Y}_\tau$ \citep{li2025back}, rather than to predict the injected noise or another noised parameterization. We minimize the following reconstruction loss across noise levels $\tau$:
\begin{equation}\mathcal{L}_{\text{forward}}(\theta) = \mathbb{E}_{\tau} \left[ -\log P_{\theta}(Y \mid \Phi_{\theta}(\tilde{Y}_\tau, X, \tau)) \right].
\end{equation}

\subsubsection{Backward-Training Models}
Backward-training models (like TRM) are trained starting from pure noise
and unrolling the recursion. Over long enough horizons, these methods must use some form of \textit{Truncated Backpropagation Through Time} (TBPTT). The common instantiation of this, as is done in TRM, is to partition the full trajectory $T$ into intervals of size $k$. The gradients are computed within a window, but the recurrent state is detached from the computation graph at the boundaries. Formally, let $\text{sg}(\cdot)$ denote the stop-gradient operator. The state update is:
\begin{align}
\tilde{H}_{t} &= \begin{cases}
\text{sg}(H_{t}) & \text{if } t\bmod k = 0 \\
H_{t} & \text{otherwise}
\end{cases}
\end{align}
The recursion continues as $H_{t+1} = \Phi_\theta(\tilde H_t, X, \dots)$. Supervision is applied at the end of each interval $k$ up to the maximum depth $T$:
\begin{equation}\label{eq:recur_loss}
\mathcal{L}_{\text{backward}}(\theta) = \sum_{i=1}^{T/k} -\log P_{\theta}(Y \mid \tilde{H}_{i \cdot k}).
\end{equation}
Unlike diffusion, the state $\tilde{H}_{i \cdot k}$ is the result of applying $\Phi_\theta$ sequentially, allowing the model to learn forward-looking algorithmic operations while maintaining tractable training dynamics via the stop-gradient operator.

\subsection{Denoising Recursion Models (DRMs)}

DRMs are a hybrid between forward-training and backward-training looped transformers. We leverage the ``fast-forward'' capability of diffusion to initialize the model at an arbitrary difficulty level (noise scale $\tau$), but we then allow the model to resolve this state over a recursive window of $k$ steps. We initialize the embedded corrupted target: $H_0 = \text{Embed}(\tilde{Y}_\tau)$
where $\tilde{Y}_\tau$ is sampled from a forward diffusion process $q_\tau(\tilde{Y}\mid Y)$. We then unroll the transition operator $\Phi_\theta$ for a fixed window of $k$ iterations and minimize the negative log-likelihood of the clean ground truth $Y$ given the resulting state $H_k$:
\begin{equation}
    \mathcal{L}_{\text{DRM}}(\theta) = \mathbb{E}_{\tau} \left[ -\log P_{\theta}(Y \mid H_k) \right].
\end{equation}
By initializing with $\tilde{Y}_\tau$, we retain the state diversity of diffusion, exposing the model to a vast range of intermediate states without needing to generate them. And by backpropagating through $k$ steps of $\Phi_\theta$, we reintroduce the forward-looking inductive bias of recurrent models, training the network to perform multi-step algorithmic refinement rather than greedy denoising. 

See Figure \ref{fig:model_arch} for an illustrative comparison. At inference, DRMs initialize from a sample at the maximally corrupted state supported by the forward process and then iteratively apply the denoiser over a sequence of corruption levels $\tau_T > \tau_{T-1} > \cdots > \tau_0 = 0$, producing an updated prediction at each step until reaching the uncorrupted state. In the masked discrete case, this corresponds to initializing all output tokens as \texttt{MASK} and repeatedly predicting a full grid followed by partial
re-masking.

\subsection{State Perturbation Recursion Models (SPRMs)}\label{sec:sprm}

DRMs start with a forward step, followed by backward steps. We consider an additional hybrid method with \textit{alternating} backward and forward processes. We call this State Perturbation Recursion Models (SPRMs) because it forces the model to continuously recover from stochastic perturbations along states of its own reasoning path. Rather than denoising from a noisy ground truth, SPRMs effectively denoise from its own incumbent solution. This increases the diversity of intermediate states, which could improve robustness at test time when encountering out-of-distribution states.

Let $q_\tau(\tilde{H}\mid H)$ denote a forward corruption process.
At each recursion step $t$, we sample a noise level $\tau_t$ and perturb the previous state via the forward process:
\begin{equation}
\tau_t \sim \mathcal{U}(1,\dots,T), \qquad 
\tilde{H}_{t-1} \sim q_{\tau_t}(\cdot \mid H_{t-1}), \qquad
H_t = \Phi_\theta(\tilde{H}_{t-1}, X).
\end{equation}
This corruption forces the operator $\Phi_\theta$ to map from the noisy latent manifold back to the solution manifold at every iteration. We think continuous state perturbations could discourage the model to be robust attractor field around correct reasoning trajectories and avoid  converging to brittle trajectories.

\section{Experimental Setup}

\subsection{Data}

We evaluate on two datasets: ARC2-Public Eval (120 difficult tasks) and ARC2-Public Train, which we refer to as \textit{ARC-Easy} (1000 generally easier tasks). For pretraining, we use three settings of increasing scale: ARC-Easy + ConceptARC \citep{moskvichev2023conceptarcbenchmarkevaluatingunderstanding}, reARC \citep{hodel2024addressing}, and NVARC-Train \citep{sorokin2025nvarc}. We exclude NVARC-Eval because it is derived from ARC2-Public Eval and would introduce data leakage. Dataset sizes are summarized in Table~\ref{tab:arc_results}.
Each ARC task provides a small set of input-output training examples and 1--3 held-out test inputs, where each grid is at most \(30 \times 30\) and each cell takes one of 10 colours. See Appendix~\ref{app:maze_sudoku} for an evaluation on Sudoku and Maze tasks.

\subsection{Model Architecture}

The methods in Section~\ref{sec:methods} are agnostic to the particular architecture used to instantiate them. In experiments, we fix a single concrete model instantiation across all methods: we use the TRM architecture and codebase of \citet{jolicoeur2025less} for the backward-training model, and we build DRM, SPRM, and all relevant baselines within this same implementation. Our comparison does not vary the transition operator $\Phi_\theta$ and the entire surrounding architectural and training scaffold inherited from TRM. Aside from modifications that are intrinsic to each training method, all components and hyperparameters are held constant. See Table~\ref{tab:model-hparams} in the appendix for the full hyperparameter details.

The TRM instantiation includes several components that were omitted from the abstract formulation in Section~\ref{sec:methods}. First, the model uses a learned \textbf{task embedding}, concatenated at each recursion step, to share information across examples from the same task. Second, the recurrent state is organized as a \textbf{scratchpad} $(y_t, z_t)$, where $y_t$ is mapped to output tokens and $z_t$ is an unconstrained latent workspace for computation. Third, the TRM training recipe includes \textbf{warm-up recursion steps} before gradients are applied. Finally, the model includes a scalar \textbf{Q-head} that predicts exact-match correctness and is used for early exit and for selecting a final solution across augmentations. See Figure \ref{fig:pseudocode_comparison}a in the appendix for pseudocode. These components are retained across our model variants unless a change is required by the method definition itself.

\subsection{DRM Corruption Process Instantiation} \label{sec:drm_implementation}
Given that we tokenize each cell in an ARC grid, we instantiate $q_\tau(\tilde{Y}\mid Y)$ as a \textit{discrete masking schedule}---an established technique in language diffusion \citep{nie2025llada}---rather than adding Gaussian noise to continuous embeddings\footnote{We first tried standard DDPM on the grid embeddings but observed very poor performance. We suspect this was because the model strategically learned embeddings that were recoverable from noise.}
. We used a cosine noise schedule, observing faster training convergence compared to a linear or sigmoid schedule (see Appendix \ref{app:noise_schedule_ablation}). This aligns with prior work on discrete diffusion \citep{chang2022maskgit} and continuous diffusion \citep{nichol2021improved}. Linear schedules tend to destroy information too often, while cosine schedules smooth out the rate of information destruction.
Let $\tau \in [0,1]$ represent the normalized diffusion timestep. We compute the signal retention rate $\bar{\alpha}_\tau$ as: 
\begin{equation}
\bar{\alpha}_\tau = \cos^2\left(\frac{\pi s}{2}\right),
\end{equation} where the fraction of tokens to be masked is $r(\tau)=1-\bar{\alpha}_\tau$. For a diffusion timestep $\tau \sim \mathcal{U}[0, 1]$, we selected $\lfloor r(\tau) \cdot M \rfloor$ indices uniformly at random; this creates a ``salt and pepper" masking that we thought could be quite easy to interpolate, however we did not observe any benefit from sampling in a spatially correlated way by using a 2D Gaussian kernel where full ``objects" were more likely to be masked. We replaced the tokens at those indices with a special \texttt{MASK} token and used separate token embeddings for mapping the masked ground truth into latent space than those used for input grids. Since TRM adds the input embedding to the output embedding, we hypothesized these separate embeddings would help to avoid interference. See Figure \ref{fig:pseudocode_comparison}b for pseudocode.

During inference, we reverse this process using an iterative ``generate-and-remask" strategy as in \citet{nie2025llada} except we do not freeze tokens in place if they are not selected to be remasked. We define a sequence of discrete timesteps $s_T, \dots, s_0$ decreasing from $1$ to $0$---for every example, we sample uniformly between 1 and 0 and sort in descending order. We initialize the grid as fully masked. At each iteration $i$, the model predicts logits for all grid cells given the current input. We first decode a fully populated candidate grid $\hat{y}_0$ by taking the argmax of the logits at every position. To construct the input for the subsequent step $i-1$, we calculate the required number of masked tokens $j = \lfloor r_{s_{i-1}} \cdot M \rfloor$ based on the schedule. We then select $j$ indices \textit{uniformly at random} to be re-masked as in training, retaining the predicted values from $\hat{y}_0$ at the unselected indices. This process of predicting the full grid and stochastically remasking a subset of cells, repeats until $s_0=0$, at which point no masks remain. We tried confidence-based remasking---i.e., masking lowest confidence tokens \citep{nie2025llada}---but did not observe any performance improvement. See Appendix \ref{app:drm_vis} for examples of DRM inference.

\subsection{SPRM Corruption Process Instantiation}

For our secondary hybrid method, we inject noise directly into the hidden states rather than the input tokens. This perturbation is applied exclusively at the truncation points (``outer loops" of TRM). We instantiate $q_\tau(\tilde{Y}\mid Y)$ with the standard linear variance schedule from Denoising Diffusion Probabilistic Models (DDPM) \citep{ho2020denoising}. We define a variance schedule $\beta_t$ that increases linearly from $\beta_\text{start}=10^{-4}$ to $\beta_\text{end}=0.02$. At the end of each truncated backpropagation window (every $k$ steps), we perturb the hidden state $H_{i\cdot k}$ before passing it to the next window. We use the variance-preserving update:
\begin{equation}
H_{\text{next}} = \sqrt{1 - \beta_\tau} H_{\text{prev}} + \sqrt{\beta_\tau} \epsilon, \quad \epsilon \sim \mathcal{N}(\mu(H_{\text{prev}}),\Sigma(H_{\text{prev}})), \tau \sim \mathcal{U}(1 \dots T).
\end{equation}
The noise is scaled with the mean and variance of the latent state $H_{\text{prev}}$. We do not add noise during inference. We tried adding noise during inference according to the same DDPM noise schedule used during training and observed minor performance reduction. 

\subsection{Baselines}

\textbf{Standard Diffusion and Transformer Baselines.} To isolate the necessity of the recursive inductive bias, we evaluated standard architectures that match the computational budget of the TRM but lack the recurrent weight sharing. We evaluated standard diffusion models using the exact same discrete masking schedule and inference procedure as the DRM implementation described in Section \ref{sec:drm_implementation}. The 7M model matches the capacity of the DRM transition operator. The 70M model is constructed by stacking transformer blocks. to match the total computational depth (FLOPs) used by the TRM during a single backpropagation window (which consists of four recurrent steps plus one final processing loop). We also evaluated a standard Transformer with the same stacked architecture. To determine if recursive layers are required specifically within a gradient block, we also evaluated the repeated application of a stacked transformer. 

\textbf{TRM without deep supervision.} We include a TRM baseline where we trained solely on the loss at the final step of the recursion window, removing the intermediate truncated backpropagation losses. This serves to measure the dependence of the architecture on dense supervision signals.

\textbf{State-of-the-Art (NVARC).} We compare our results against the current state-of-the-art (among open-source models) NVARC system \citep{sorokin2025nvarc}, which achieved first place in the 2025 ARC Challenge. We evaluated their 4-billion parameter LLM while controlling for the dataset.

\textbf{Other baselines} We also compare to ViARC \cite{hu2025arc} with reARC pretraining data. We observed severe optimization instability when pretraining on NVARC data so we do not include those results. We were not able to benchmark URM \citep{gao2025universal} because of persistent training crashes.

\subsection[Training and Evaluation]{Training and Evaluation}
We follow the standard TRM protocol of \citet{jolicoeur2025less} unless specified otherwise. We did all exploratory hyperparameter tuning on ARC2-Easy to avoid overfitting ARC2-Eval. See the appendix for specific training (Table \ref{tab:training-hparams}) and architecture (Table \ref{tab:model-hparams}) parameters. See our official GitHub repository \href{https://github.com/wwwwwwwwz/DenoisingRecursionModels}{here}. 

\textbf{Data Preparation and Augmentation.} For each ARC task, we sample 1000 augmentations. For each augmentation we (1) randomly permute the colours, (2) select a random transformation (e.g., flip, rotate) and (3) if the grid is smaller than 30$\times$30, randomly translate each pair of grids into a fixed 30$\times$30 template (cells outside the original grid are masked when loss is measured). We flatten grids in row-major order with learned colour/\texttt{MASK} embeddings and RoPE \citep{su2024roformer}.

\textbf{Evaluation Protocol.} We follow the released TRM implementation exactly. Although the TRM paper describes ARC evaluation as selecting the most common prediction over 1000 test-time augmentations, the public code additionally aggregates predictions across evaluation snapshots during finetuning \citep{huang2017snapshot}. We used this protocol for a direct comparison to TRM. In our setup, evaluations were performed every 10K epochs, so the final candidate pool contains predictions from all saved finetuning checkpoints and all 1000 augmentations. After undoing the augmentations, identical predictions are grouped across this full pool and primarily by vote count, with mean predicted $q$ used as a tie-breaker---Table \ref{tab:arc_results} reports pass@2 exact-match accuracy.  In our own ablations, evaluating only the final checkpoint tended to give slightly worse results.

\textbf{Pretraining and Finetuning.} For the large-scale data settings, we used a two-stage process: we pretrained the model on reARC for 100K epochs and on NVARC+reARC for 2K epochs, using identical hardware for both. Following pretraining, we performed a finetuning stage of 300K epochs, where we strictly followed the TRM methodology of jointly training on the provided training pairs of both ARC2-Train and ARC2-Eval tasks. We save checkpoints every 10K finetuning epochs, and final ARC2-Eval predictions are obtained by aggregating candidates from all saved checkpoints using the $q$-value ranking procedure described above, rather than evaluating only the terminal checkpoint.

\textbf{NVARC Baseline Details.} For the NVARC baseline, we pretrained until the validation loss plateaued (approx. 2K epochs). We used \citet{sorokin2025nvarc}'s test-time training implementation for finetuning to ARC2-Easy and ARC2-Eval.

\section{Results}\label{section:results}

\begin{table*}[t]
    \centering
    \caption{Pass@2 exact-match accuracy results on held-out data for every task comparing \textbf{DRM} and \textbf{SPRM)} methods against the TRM baseline and state-of-the-art NVARC. The 7M models use identical architectures. 
    The 14M settings increased embedding size from 512 to 768, number of heads from 8 to 12, and gradient loops $k$ from 4 to 6. *indicates significant improvement over TRM (permutation test, $p<.05$) with identical model architecture ($p$-values in Appendix \ref{app:significance_results}).
    }
    \label{tab:arc_results}
    
    \setlength{\tabcolsep}{1pt}
    
    \begin{tabular}{l c c c c c}
        \toprule
        \multirow{3}{*}{\textbf{Method}} & \multirow{3}{*}{\textbf{\#Params}} & \textbf{ARC-Easy} & \multicolumn{3}{c}{\textbf{ARC2-Eval (by Pretraining Data)}} \\
        \cmidrule(lr){4-6}
         & & \scriptsize{(No Pretrain)} & \textbf{+ARC-Easy} & \textbf{+reARC} & \textbf{+NVARC Train} \\
         & & \scriptsize\textit{-} & \scriptsize\textit{1k tasks (3.5 ex/task)} & \scriptsize\textit{400 tasks (1k ex/task)} & \scriptsize\textit{47k tasks (24 ex/task)} \\
        \midrule
        
        \multicolumn{6}{l}{\textit{\textbf{Baselines}}} \\
        Transformer & 7M & 10.9 & - & - & - \\
        Transformer & 70M & 25.5 & 1.7 & - & - \\
        Transformer \textit{w/ deep sup} & 70M & 37.2 & 2.5 & - & - \\
        Masked Diffusion (Standard) & 7M & 0.0 & 0.0 & - & - \\
        Masked Diffusion (Standard) & 70M & 0.7 & 0.0 & - & - \\
        TRM \textit{w/o deep sup} & 7M & 34.7 & 3.3 & - & - \\
        TRM \textit{w/o inner loop} & 34M & 39.8 & 3.0 & - & - \\
        ViARC & 18M & - & - & 8.3 & - \\
        LLM (NVARC) & 4B & 43.2 & \colorbox{green!30}{\textbf{13.8}} & 15.6 & 22.6 \\
        \midrule
        TRM (Baseline) & \multirow{3}{*}{7M} & 45.7 & 6.3 & 12.4 & 12.5 \\
        SPRM (Ours) & & \colorbox{green!30}{\textbf{50.7}$^{*}$} & \colorbox{green!30}{\textbf{10.1}$^{*}$} & 11.8 & 14.4 \\
        \textbf{DRM (Ours)} & & 50.5$^{*}$ & 9.6$^{*}$ & \colorbox{green!30}{\textbf{14.7}} & \colorbox{green!30}{\textbf{16.7}} \\
        \midrule
        TRM (Baseline) & \multirow{3}{*}{14M} & 53.2 & 10.6 & 18.8 & 21.8 \\
        SPRM (Ours) & & \colorbox{green!30}{\textbf{55.4}$^{*}$} & 11.8 & 17.4 & 22.2 \\
        \textbf{DRM (Ours)} & & 55.0$^{*}$ & \colorbox{green!30}{\textbf{13.3}} & \colorbox{green!30}{\textbf{21.0}} & \colorbox{green!30}{\textbf{24.9}} \\
        \bottomrule
    \end{tabular}
\end{table*}

\textbf{The Necessity of Recursion.} We first establish the baseline difficulty of the task. We find that standard non-recursive architectures are insufficient for ARC-AGI. A standard transformer achieved only 25.5\% on ARC-Easy, far below the recursive TRM baseline of 45.7\%. Crucially, we found that diffusion without backpropagating through multiple recursive steps performs very poorly. A standard diffusion model (matching the DRM architecture but without looping) failed completely, achieving $\sim$0\% accuracy. Even when we stacked layers to match the compute budget of recursion (70M params), we still could not get above 1\%.  This suggests that iterative refinement via shared weights is the primary driver of performance.

\textbf{DRM Stabilizes Training without Deep Supervision.} Standard TRMs rely heavily on deep supervision (TBPTT) to guide intermediate states. When we removed this supervision (``TRM w/o deep sup"), performance fell by 11\% (from 45.7\% to 34.7\%) on ARC-Easy. In contrast, our DRM method trains without any intermediate gradient truncation or deep supervision yet outperformed the TRM baseline by nearly 5\% on ARC-Easy (50.5\% vs 45.7\%). This suggests that the diffusion objective provides a useful curriculum for learning long-horizon reasoning trajectories.

\textbf{DRM Outperforms TRM.} These gains transfer to the ARC2-Eval benchmark, which consists of tasks with higher difficulty. On the base setting (no pretraining), DRM boosted ARC2-Eval performance from 6.3\%\footnote{\citet{jolicoeur2025less} reported 7.8\% in their paper with the identical setup and codebase.} to 9.6\% and the performance of TRM without deep supervision was just 3.3\%. Even in the high-data regime (NVARC pretraining), DRM improved ARC2-Eval accuracy from 12.5\% to 16.7\% (7M params) and 21.8\% to 24.9\% (14M params).

\textbf{Performance Scaled with Data and Model Size.}
We observe a strong synergistic effect between model and dataset size. Scaling from the 7 million parameter baseline (and four recurrent gradients steps) to the 14 million parameter (and six recurrent gradients steps) improved performance 3--4\% when restricted to pretraining on ARC2-Train. The increased capacity from a bigger model became more critical when leveraging massive pretraining datasets. Adding NVARC to pretraining improved 7M DRM by approximately 7\% but with 14 million parameters, this boosted performance by 11.5\%. This indicates that the smaller model is capacity constrained, whereas the 14M architecture can absorb the high diversity of the 47K tasks from NVARC.

\textbf{DRM Outperformed SOTA 4B LLM Method.} Our 14M-parameter DRM outperformed the current state-of-the-art NVARC (Qwen3-4B Base) model (24.9\% vs 22.6\%) when trained on the same data. This result is particularly striking because NVARC is a highly engineered system built around a 4-billion parameter model, specialized specifically for ARC-AGI. This demonstrated that DRM (and TRM) can be orders of magnitude more parameter efficient (14M vs. 4B).

\textbf{State Perturbation Recursion Model.} Our SPRM method also improved over the TRM baseline (50.7\% vs. 45.7\% on ARC2-Train and 6.3\% vs. 10.1\% on ARC2-Eval ). However, its effectiveness was mixed in other settings. On the reARC dataset, SPRM performed slightly worse than the TRM baseline. We hypothesize that the high diversity of reARC (1000 examples per task) saturated the model's need for exploration, making noise injection redundant.

\section{Discussion}

\textbf{ARC vs. other settings where diffusion excels.} Our results highlight the particular type of iterative refinement needed for ARC-style algorithmic reasoning. Discrete masked diffusion baselines fail almost completely even when scaled to match the TRM compute budget. ARC is a discriminative point-to-point mapping rather than to a distribution-to-distribution generative setting where diffusion usually excels. In image generation, denoising can often progress from coarse global structure to fine local detail, and many outputs may be acceptable. ARC has a single exact target, and many tasks require globally coordinated discrete transformations where locally plausible partial states are incompatible with the final solution. This mismatch may help explain why standard masked diffusion performs poorly even though diffusion is highly effective in conventional generative domains.

\textbf{Importance of Backpropagating Through Recursion.}
While DRMs substantially improve over TRMs, the difference in performance is much more modest than the gap between TRM and diffusion baselines. These findings suggest that the core ingredient to the success of these small models is backpropagating through a window of recursions. We see two complementary mechanisms. First, this provides \textit{multi-step credit assignment}: intermediate states are trained to be useful because they enable progress several steps later, which discourages greedy updates that could be very poor for certain tasks---e.g., in obstacle navigation, a move that decreases Manhattan distance can still result in a dead end. Recursive denoising creates conditions under which planning-like latent computation \citep{guez2019investigation} could emerge. Second, weight tying across depth can act as a strong regularizer for algorithmic reasoning. Because the same parameters must operate across many latent states and depths, gradients from different unrolled steps must largely agree for an update to persist; spurious, step-specific correlations are more likely to generate conflicting signals that cancel, biasing learning toward stable, reusable computations.

\textbf{DRM balances ``off-policy'' and ``on-policy'' learning.}
DRM improves upon TRM in part by mitigating long-horizon training instability. Pure backward-training is effectively ``on-policy'': the model must discover long trajectories from noise using only its own self-generated intermediate states, which are often uninformative early in training. 
DRM initializes training from diffusion-corrupted targets, providing an ``off-policy'' curriculum of easier intermediate states closer to the solution. From that starting point onward, training is explicitly ``on-policy'' where each later state is generated by the model itself rather than supplied by the corruption process. This preserves the forward-looking behavior and train--test alignment that make TRM effective.

\textbf{Scaling Behaviour.}
The strong interaction between model size, unrolled gradient depth, and pretraining scale suggest that recursion models are not limited to the ``tiny" regime, rather their parameter efficiency can be converted into improved performance as data and compute increase. 

\textbf{SOTA commercial LLMs.}
Although our method outperformed the strongest commercial LLMs available in November 2025, the frontier has advanced rapidly since then. Recent systems such as Gemini 3 Deep Think \citep{gemini_deepthink_2026} and GPT-5.4 Pro (xHigh) \citep{openai_gpt54_pro_xhigh_2026} exceed 80\% on this benchmark, far above our best result of 25\%. Direct comparison should be interpreted cautiously: the architectures and training pipelines of these systems are not public, but they surely use orders of magnitude more data and compute than our models, and may also incorporate benchmark-targeted synthetic data. In addition, agentic systems such as Poetiq's \citep{poetiq_meta_system_2025} suggest that coding-based search and verification loops are especially effective for ARC-AGI. We suspect that such mechanisms play an important role in these recent gains. We evaluated our model at just 7M or 14M parameters, where learning general coding capabilities is likely out of reach.

\section{Conclusions}
We introduced the Denoising Recursion Model (DRM), a looped transformer that resolves the opposing limitations of (1) backward-training methods that struggle with long-horizon stability and (2) diffusion that suffers from a myopic objective and a misalignment between training (on noised target) and testing (on self-generated trajectories). We effectively combine the training objectives: we corrupt the target with noise identical to the diffusion process, but train the model to recover the clean signal over a sequence of recursive steps, rather than a single jump. Our method consistently outperformed the TRM baseline on ARC-AGI across a variety of settings. On the largest pretraining dataset, we achieved 24.9\% accuracy on ARC2-Eval, surpassing the winner of the recent ARC Challenge (NVARC) when trained on identical data. We also showed large improvements from scaling the architecture from 7M to 14M parameters, suggesting that further scaling could yield substantial gains. At present, progress on ARC-AGI is difficult to interpret because gains from model architecture and training objectives are often entangled with gains from more sophisticated synthetic-data generation. Our results suggest that substantial headroom remains in the learning algorithm itself, and motivate evaluations that make these sources of progress easier to disentangle.

\newpage

\bibliographystyle{apalike}
\bibliography{diffusion_recursion}

@inproceedings{nie2025llada,
  title={Large Language Diffusion Models},
  author={Nie, Shen and Zhu, Fengqi and You, Zebin and Zhang, Xiaolu and Ou, Jingyang and Hu, Jun and ZHOU, JUN and Lin, Yankai and Wen, Ji-Rong and Li, Chongxuan},
  booktitle={The Thirty-ninth Annual Conference on Neural Information Processing Systems},
  year={2025}
}

@article{lai2025principles,
  title={The principles of diffusion models},
  author={Lai, Chieh-Hsin and Song, Yang and Kim, Dongjun and Mitsufuji, Yuki and Ermon, Stefano},
  journal={arXiv preprint arXiv:2510.21890},
  year={2025}
}

@misc{openaio3,
  title = {{Introducing OpenAI o3 and o4-mini}},
  author={OpenAI},
  howpublished = {\url{https://openai.com/index/introducing-o3-and-o4-mini/}},
  year={2025},
}

@article{franzen2025product,
  title={Product of Experts with LLMs: Boosting Performance on ARC Is a Matter of Perspective},
  author={Franzen, Daniel and Disselhoff, Jan and Hartmann, David},
  journal={arXiv preprint arXiv:2505.07859},
  year={2025}
}

@article{chollet2025arc,
  title={Arc-agi-2: A new challenge for frontier ai reasoning systems},
  author={Chollet, Francois and Knoop, Mike and Kamradt, Gregory and Landers, Bryan and Pinkard, Henry},
  journal={arXiv preprint arXiv:2505.11831},
  year={2025}
}

@article{pourcel2025self,
  title={Self-Improving Language Models for Evolutionary Program Synthesis: A Case Study on ARC-AGI},
  author={Pourcel, Julien and Colas, C{\'e}dric and Oudeyer, Pierre-Yves},
  journal={arXiv preprint arXiv:2507.14172},
  year={2025}
}

@article{akyurek2024surprising,
  title={The surprising effectiveness of test-time training for few-shot learning},
  author={Aky{\"u}rek, Ekin and Damani, Mehul and Zweiger, Adam and Qiu, Linlu and Guo, Han and Pari, Jyothish and Kim, Yoon and Andreas, Jacob},
  journal={arXiv preprint arXiv:2411.07279},
  year={2024}
}

@article{hodel2024addressing,
  title={Addressing the abstraction and reasoning corpus via procedural example generation},
  author={Hodel, Michael},
  journal={arXiv preprint arXiv:2404.07353},
  year={2024}
}

@inproceedings{xu2023graphs,
  title={Graphs, constraints, and search for the abstraction and reasoning corpus},
  author={Xu, Yudong and Khalil, Elias B and Sanner, Scott},
  booktitle={Proceedings of the AAAI Conference on Artificial Intelligence},
  volume={37},
  number={4},
  pages={4115--4122},
  year={2023}
}

@article{wang2025hierarchical,
  title={Hierarchical Reasoning Model},
  author={Wang, Guan and Li, Jin and Sun, Yuhao and Chen, Xing and Liu, Changling and Wu, Yue and Lu, Meng and Song, Sen and Yadkori, Yasin Abbasi},
  journal={arXiv preprint arXiv:2506.21734},
  year={2025}
}

@article{cole2025don,
  title={Don't throw the baby out with the bathwater: How and why deep learning for ARC},
  author={Cole, Jack and Osman, Mohamed},
  journal={arXiv preprint arXiv:2506.14276},
  year={2025}
}

@article{you2025llada,
  title={Llada-v: Large language diffusion models with visual instruction tuning},
  author={You, Zebin and Nie, Shen and Zhang, Xiaolu and Hu, Jun and Zhou, Jun and Lu, Zhiwu and Wen, Ji-Rong and Li, Chongxuan},
  journal={arXiv preprint arXiv:2505.16933},
  year={2025}
}

@article{jolicoeur2025less,
  title={Less is More: Recursive Reasoning with Tiny Networks},
  author={Jolicoeur-Martineau, Alexia},
  journal={arXiv preprint arXiv:2510.04871},
  year={2025}
}

@article{chung2015recurrent,
  title={A recurrent latent variable model for sequential data},
  author={Chung, Junyoung and Kastner, Kyle and Dinh, Laurent and Goel, Kratarth and Courville, Aaron C and Bengio, Yoshua},
  journal={Advances in neural information processing systems},
  volume={28},
  year={2015}
}

@article{geiping2025efficient,
  title={Efficient Parallel Samplers for Recurrent-Depth Models and Their Connection to Diffusion Language Models},
  author={Geiping, Jonas and Yang, Xinyu and Su, Guinan},
  journal={arXiv preprint arXiv:2510.14961},
  year={2025}
}

@article{geiping2025scaling,
  title={Scaling up test-time compute with latent reasoning: A recurrent depth approach},
  author={Geiping, Jonas and McLeish, Sean and Jain, Neel and Kirchenbauer, John and Singh, Siddharth and Bartoldson, Brian R and Kailkhura, Bhavya and Bhatele, Abhinav and Goldstein, Tom},
  journal={arXiv preprint arXiv:2502.05171},
  year={2025}
}

@article{lim2021noisy,
  title={Noisy recurrent neural networks},
  author={Lim, Soon Hoe and Erichson, N Benjamin and Hodgkinson, Liam and Mahoney, Michael W},
  journal={Advances in Neural Information Processing Systems},
  volume={34},
  pages={5124--5137},
  year={2021}
}

@inproceedings{dieng2018noisin,
  title={Noisin: Unbiased regularization for recurrent neural networks},
  author={Dieng, Adji Bousso and Ranganath, Rajesh and Altosaar, Jaan and Blei, David},
  booktitle={International Conference on Machine Learning},
  pages={1252--1261},
  year={2018},
  organization={PMLR}
}

@article{chen2024diffusionforcingnexttokenprediction,
  title={Diffusion forcing: Next-token prediction meets full-sequence diffusion},
  author={Chen, Boyuan and Mart{\'\i} Mons{\'o}, Diego and Du, Yilun and Simchowitz, Max and Tedrake, Russ and Sitzmann, Vincent},
  journal={Advances in Neural Information Processing Systems},
  volume={37},
  pages={24081--24125},
  year={2024}
}

@article{merrill2025little,
  title={A little depth goes a long way: The expressive power of log-depth transformers},
  author={Merrill, William and Sabharwal, Ashish},
  journal={arXiv preprint arXiv:2503.03961},
  year={2025}
}

@misc{zhou2025coevolutionarycontinuousdiscretediffusion,
      title={Coevolutionary Continuous Discrete Diffusion: Make Your Diffusion Language Model a Latent Reasoner}, 
      author={Cai Zhou and Chenxiao Yang and Yi Hu and Chenyu Wang and Chubin Zhang and Muhan Zhang and Lester Mackey and Tommi Jaakkola and Stephen Bates and Dinghuai Zhang},
      year={2025},
      eprint={2510.03206},
      archivePrefix={arXiv},
      primaryClass={cs.AI},
      url={https://arxiv.org/abs/2510.03206}, 
}

@article{dehghani2018universal,
  title={Universal transformers},
  author={Dehghani, Mostafa and Gouws, Stephan and Vinyals, Oriol and Uszkoreit, Jakob and Kaiser, {\L}ukasz},
  journal={arXiv preprint arXiv:1807.03819},
  year={2018}
}

@article{ho2020denoising,
  title={Denoising diffusion probabilistic models},
  author={Ho, Jonathan and Jain, Ajay and Abbeel, Pieter},
  journal={Advances in neural information processing systems},
  volume={33},
  pages={6840--6851},
  year={2020}
}

@article{zhu2025llada,
  title={LLaDA 1.5: Variance-Reduced Preference Optimization for Large Language Diffusion Models},
  author={Zhu, Fengqi and Wang, Rongzhen and Nie, Shen and Zhang, Xiaolu and Wu, Chunwei and Hu, Jun and Zhou, Jun and Chen, Jianfei and Lin, Yankai and Wen, Ji-Rong and others},
  journal={arXiv preprint arXiv:2505.19223},
  year={2025}
}

@incollection{williams1995gradient,
  title={Gradient-based learning algorithms for recurrent networks and their computational complexity},
  author={Williams, Ronald J and Zipser, David},
  booktitle={Backpropagation: theory, architectures, and applications},
  pages={433--486},
  year={1995}
}

@inproceedings{giannou2023looped,
  title={Looped transformers as programmable computers},
  author={Giannou, Angeliki and Rajput, Shashank and Sohn, Jy-yong and Lee, Kangwook and Lee, Jason D and Papailiopoulos, Dimitris},
  booktitle={International Conference on Machine Learning},
  pages={11398--11442},
  year={2023},
  organization={PMLR}
}

@inproceedings{sohl2015deep,
  title={Deep unsupervised learning using nonequilibrium thermodynamics},
  author={Sohl-Dickstein, Jascha and Weiss, Eric and Maheswaranathan, Niru and Ganguli, Surya},
  booktitle={International conference on machine learning},
  pages={2256--2265},
  year={2015},
  organization={PMLR}
}

@inproceedings{he2016deep,
  title={Deep residual learning for image recognition},
  author={He, Kaiming and Zhang, Xiangyu and Ren, Shaoqing and Sun, Jian},
  booktitle={Proceedings of the IEEE conference on computer vision and pattern recognition},
  pages={770--778},
  year={2016}
}

@article{bai2019deep,
  title={Deep equilibrium models},
  author={Bai, Shaojie and Kolter, J Zico and Koltun, Vladlen},
  journal={Advances in neural information processing systems},
  volume={32},
  year={2019}
}

@article{bai2021stabilizing,
  title={Stabilizing equilibrium models by jacobian regularization},
  author={Bai, Shaojie and Koltun, Vladlen and Kolter, J Zico},
  journal={arXiv preprint arXiv:2106.14342},
  year={2021}
}

@misc{sorokin2025nvarc,
  title  = {{NVARC} Solution to {ARC-AGI-2} 2025},
  author = {Sorokin, Ivan and Puget, Jean-François},
  year   = {2025},
  url    = {https://www.kaggle.com/competitions/arc-prize-2025/writeups/nvarc},
  note   = {Kaggle ARC Prize 2025, 1st Place Solution. Also available at \url{https://github.com/1ytic/NVARC}}
}

@article{elman1990finding,
  title={Finding structure in time},
  author={Elman, Jeffrey L},
  journal={Cognitive science},
  volume={14},
  number={2},
  pages={179--211},
  year={1990},
  publisher={Wiley Online Library}
}

@article{chollet2019measure,
  title={On the measure of intelligence},
  author={Chollet, Fran{\c{c}}ois},
  journal={arXiv preprint arXiv:1911.01547},
  year={2019}
}

@misc{Poetiq2025,
  author       = {{Poetiq Team}},
  title        = {Traversing the Frontier of Superintelligence},
  howpublished = {Poetiq AI Blog},
  year         = {2025},
  url          = {https://poetiq.ai/posts/arcagi_verified/},
  note         = {Accessed: 2026-01-26}
}

@article{franzen2025architects,
  title={The ARChitects ARC Prize 2025 Solution: Recursive Masked Diffusion for 2D Reasoning},
  author={Franzen, Daniel and Disselhoff, Jan and Hartmann, David},
  journal={ARC Prize 2025 Technical Reports},
  year={2025},
  url={https://lambdalabsml.github.io/ARC2025_Solution_by_the_ARChitects/}
}

@article{gao2025universal,
  title={Universal Reasoning Model},
  author={Gao, Zitian and Chen, Lynx and Xiao, Yihao and Xing, He and Tao, Ran and Luo, Haoming and Zhou, Joey and Dai, Bryan},
  journal={arXiv preprint arXiv:2512.14693},
  year={2025}
}

@inproceedings{nichol2021improved,
  title={Improved denoising diffusion probabilistic models},
  author={Nichol, Alexander Quinn and Dhariwal, Prafulla},
  booktitle={International conference on machine learning},
  pages={8162--8171},
  year={2021},
  organization={PMLR}
}

@article{hu2025arc,
  title={ARC Is a Vision Problem!},
  author={Hu, Keya and Cy, Ali and Qiu, Linlu and Ding, Xiaoman Delores and Wang, Runqian and Zhu, Yeyin Eva and Andreas, Jacob and He, Kaiming},
  journal={arXiv preprint arXiv:2511.14761},
  year={2025}
}

@article{austin2021structured,
  title={Structured denoising diffusion models in discrete state-spaces},
  author={Austin, Jacob and Johnson, Daniel D and Ho, Jonathan and Tarlow, Daniel and Van Den Berg, Rianne},
  journal={Advances in neural information processing systems},
  volume={34},
  pages={17981--17993},
  year={2021}
}

@inproceedings{chang2022maskgit,
  title={Maskgit: Masked generative image transformer},
  author={Chang, Huiwen and Zhang, Han and Jiang, Lu and Liu, Ce and Freeman, William T},
  booktitle={Proceedings of the IEEE/CVF conference on computer vision and pattern recognition},
  pages={11315--11325},
  year={2022}
}

@article{hochreiter1997long,
  title={Long short-term memory},
  author={Hochreiter, Sepp and Schmidhuber, J{\"u}rgen},
  journal={Neural computation},
  volume={9},
  number={8},
  pages={1735--1780},
  year={1997},
  publisher={MIT press}
}

@article{selsam2018learning,
  title={Learning a SAT solver from single-bit supervision},
  author={Selsam, Daniel and Lamm, Matthew and B{\"u}nz, Benedikt and Liang, Percy and de Moura, Leonardo and Dill, David L},
  journal={arXiv preprint arXiv:1802.03685},
  year={2018}
}

@misc{fan2025loopedtransformerslengthgeneralization,
      title={Looped Transformers for Length Generalization}, 
      author={Ying Fan and Yilun Du and Kannan Ramchandran and Kangwook Lee},
      year={2025},
      eprint={2409.15647},
      archivePrefix={arXiv},
      primaryClass={cs.LG},
      url={https://arxiv.org/abs/2409.15647}, 
}

@misc{zhu2025scalinglatentreasoninglooped,
      title={Scaling Latent Reasoning via Looped Language Models}, 
      author={Rui-Jie Zhu and Zixuan Wang and Kai Hua and Tianyu Zhang and Ziniu Li and Haoran Que and Boyi Wei and Zixin Wen and Fan Yin and He Xing and Lu Li and Jiajun Shi and Kaijing Ma and Shanda Li and Taylor Kergan and Andrew Smith and Xingwei Qu and Mude Hui and Bohong Wu and Qiyang Min and Hongzhi Huang and Xun Zhou and Wei Ye and Jiaheng Liu and Jian Yang and Yunfeng Shi and Chenghua Lin and Enduo Zhao and Tianle Cai and Ge Zhang and Wenhao Huang and Yoshua Bengio and Jason Eshraghian},
      year={2025},
      eprint={2510.25741},
      archivePrefix={arXiv},
      primaryClass={cs.CL},
      url={https://arxiv.org/abs/2510.25741}, 
}

@article{saunshi2025reasoning,
  title={Reasoning with latent thoughts: On the power of looped transformers},
  author={Saunshi, Nikunj and Dikkala, Nishanth and Li, Zhiyuan and Kumar, Sanjiv and Reddi, Sashank J},
  journal={arXiv preprint arXiv:2502.17416},
  year={2025}
}

@misc{moskvichev2023conceptarcbenchmarkevaluatingunderstanding,
      title={The ConceptARC Benchmark: Evaluating Understanding and Generalization in the ARC Domain}, 
      author={Arseny Moskvichev and Victor Vikram Odouard and Melanie Mitchell},
      year={2023},
      eprint={2305.07141},
      archivePrefix={arXiv},
      primaryClass={cs.LG},
      url={https://arxiv.org/abs/2305.07141}, 
}

@article{lee2018deterministic,
  title={Deterministic non-autoregressive neural sequence modeling by iterative refinement},
  author={Lee, Jason and Mansimov, Elman and Cho, Kyunghyun},
  journal={arXiv preprint arXiv:1802.06901},
  year={2018}
}

@article{su2024roformer,
  title={Roformer: Enhanced transformer with rotary position embedding},
  author={Su, Jianlin and Ahmed, Murtadha and Lu, Yu and Pan, Shengfeng and Bo, Wen and Liu, Yunfeng},
  journal={Neurocomputing},
  volume={568},
  pages={127063},
  year={2024},
  publisher={Elsevier}
}

@article{he2025mdpo,
  title={Mdpo: Overcoming the training-inference divide of masked diffusion language models},
  author={He, Haoyu and Renz, Katrin and Cao, Yong and Geiger, Andreas},
  journal={arXiv preprint arXiv:2508.13148},
  year={2025}
}

@inproceedings{black2024diffusionrl,
  title={Training Diffusion Models with Reinforcement Learning},
  author={Black, Kevin and Janner, Michael and Du, Yilun and Kostrikov, Ilya and Levine, Sergey},
  booktitle={The Twelfth International Conference on Learning Representations},
  year={2024}
}

@inproceedings{bush2025,
  title={Interpreting Emergent Planning in Model-Free Reinforcement Learning},
  author={Bush, Thomas and Chung, Stephen and Anwar, Usman and Garriga-Alonso, Adri{\`a} and Krueger, David},
  booktitle={The Thirteenth International Conference on Learning Representations},
  year={2025}
}

@inproceedings{guez2019investigation,
  title={An investigation of model-free planning},
  author={Guez, Arthur and Mirza, Mehdi and Gregor, Karol and Kabra, Rishabh and Racani{\`e}re, S{\'e}bastien and Weber, Th{\'e}ophane and Raposo, David and Santoro, Adam and Orseau, Laurent and Eccles, Tom and others},
  booktitle={International conference on machine learning},
  pages={2464--2473},
  year={2019},
  organization={PMLR}
}

@inproceedings{huang2017snapshot,
  title={Snapshot ensembles: Train 1, get M for free},
  author={Huang, Gao and Li, Yixuan and Pleiss, Geoff and Liu, Zhuang and Hopcroft, John E and Weinberger, Kilian Q},
  booktitle={5th International Conference on Learning Representations, ICLR 2017},
  year={2017}
}

@article{li2025back,
  title={Back to Basics: Let Denoising Generative Models Denoise},
  author={Li, Tianhong and He, Kaiming},
  journal={arXiv preprint arXiv:2511.13720},
  year={2025}
}

@software{gemini_deepthink_2026,
  author       = {{Google DeepMind}},
  title        = {Gemini Deep Think},
  year         = {2026},
  url          = {https://deepmind.google/models/gemini/},
  note         = {Large language model}
}

@software{openai_gpt54_pro_xhigh_2026,
  author       = {{OpenAI}},
  title        = {GPT-5.4 Pro XHigh},
  year         = {2026},
  url          = {https://platform.openai.com/docs/models},
  version      = {GPT-5.4 Pro XHigh}, 
  note         = {Large language model}
}

@misc{poetiq_meta_system_2025,
  author       = {{Poetiq}},
  title        = {Poetiq Meta-System and ARC-AGI Solver},
  year         = {2025},
  month        = {dec},
  howpublished = {\url{https://poetiq.ai/posts/arcagi_verified/}},
  note         = {Software wrapper. Accessed: 2026-04-17}
}

@article{gong2022diffuseq,
  title={Diffuseq: Sequence to sequence text generation with diffusion models},
  author={Gong, Shansan and Li, Mukai and Feng, Jiangtao and Wu, Zhiyong and Kong, LingPeng},
  journal={arXiv preprint arXiv:2210.08933},
  year={2022}
}

@article{li2022diffusion,
  title={Diffusion-lm improves controllable text generation},
  author={Li, Xiang and Thickstun, John and Gulrajani, Ishaan and Liang, Percy S and Hashimoto, Tatsunori B},
  journal={Advances in neural information processing systems},
  volume={35},
  pages={4328--4343},
  year={2022}
}

\appendix

\section{Background}

\subsection{ARC-AGI} \label{app:background}

\begin{figure}[h!]
    \centering
    \includegraphics[width=0.8\linewidth]{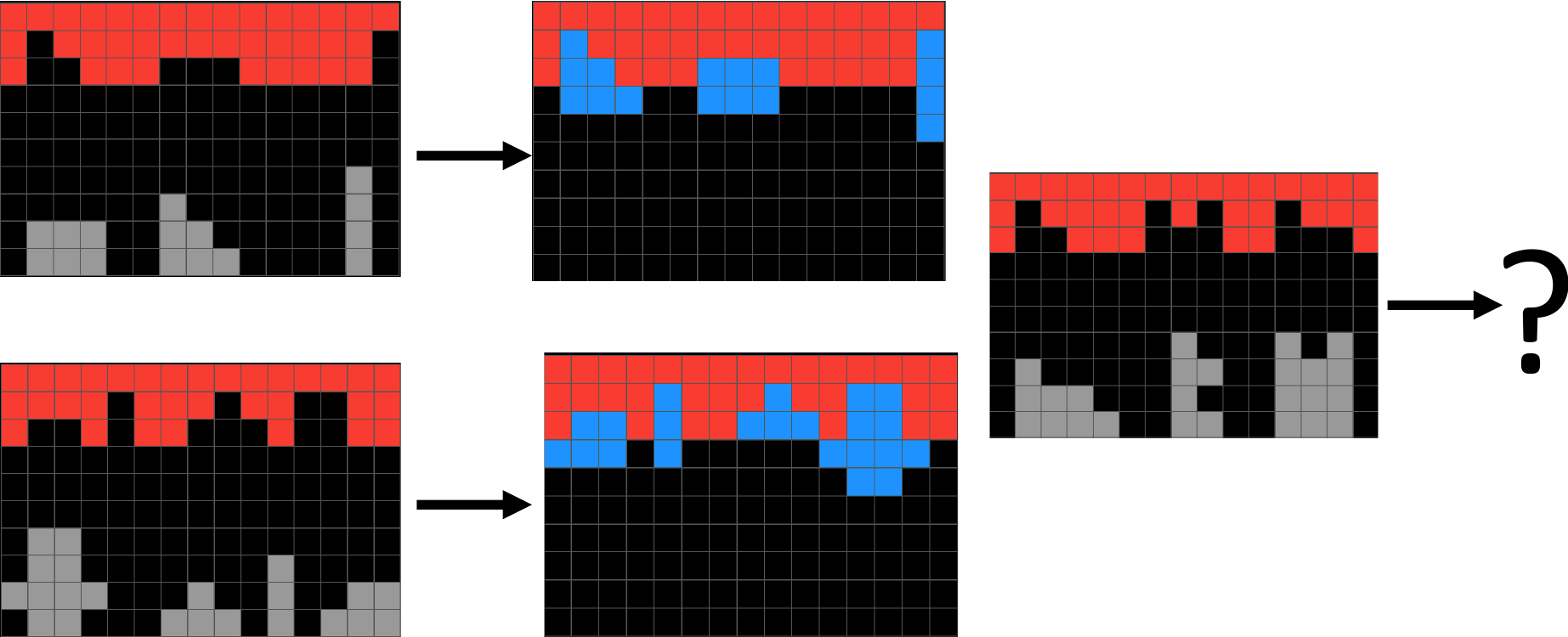}
    \caption{Example of ARC-AGI task. Two input-output grids examples for the task transformation: move each gray blocks in the input into the corresponding slots in the output and change colours to blue. Must apply the transformation correctly to a held-out input to be correct.}
    \label{fig:task_example}
\end{figure}

While modern AI systems excel at tasks where massive training data is available, they struggle with test-time adaptation---the ability to acquire new skills on the fly during deployment. Accurately evaluating this adaptability is difficult: given the scale of modern pretraining corpora, it is nearly impossible to verify whether a benchmark task is truly novel to a model or simply recalled from memory. To address this challenge, \citet{chollet2019measure} introduced the Abstraction and Reasoning Corpus (ARC-AGI): a diverse set of novel tasks where each task provides a few-shot context (2--5 input-output grid pairs). See Figure \ref{fig:task_example} for an example.

\begin{figure}
    \centering
    \includegraphics[width=\linewidth]{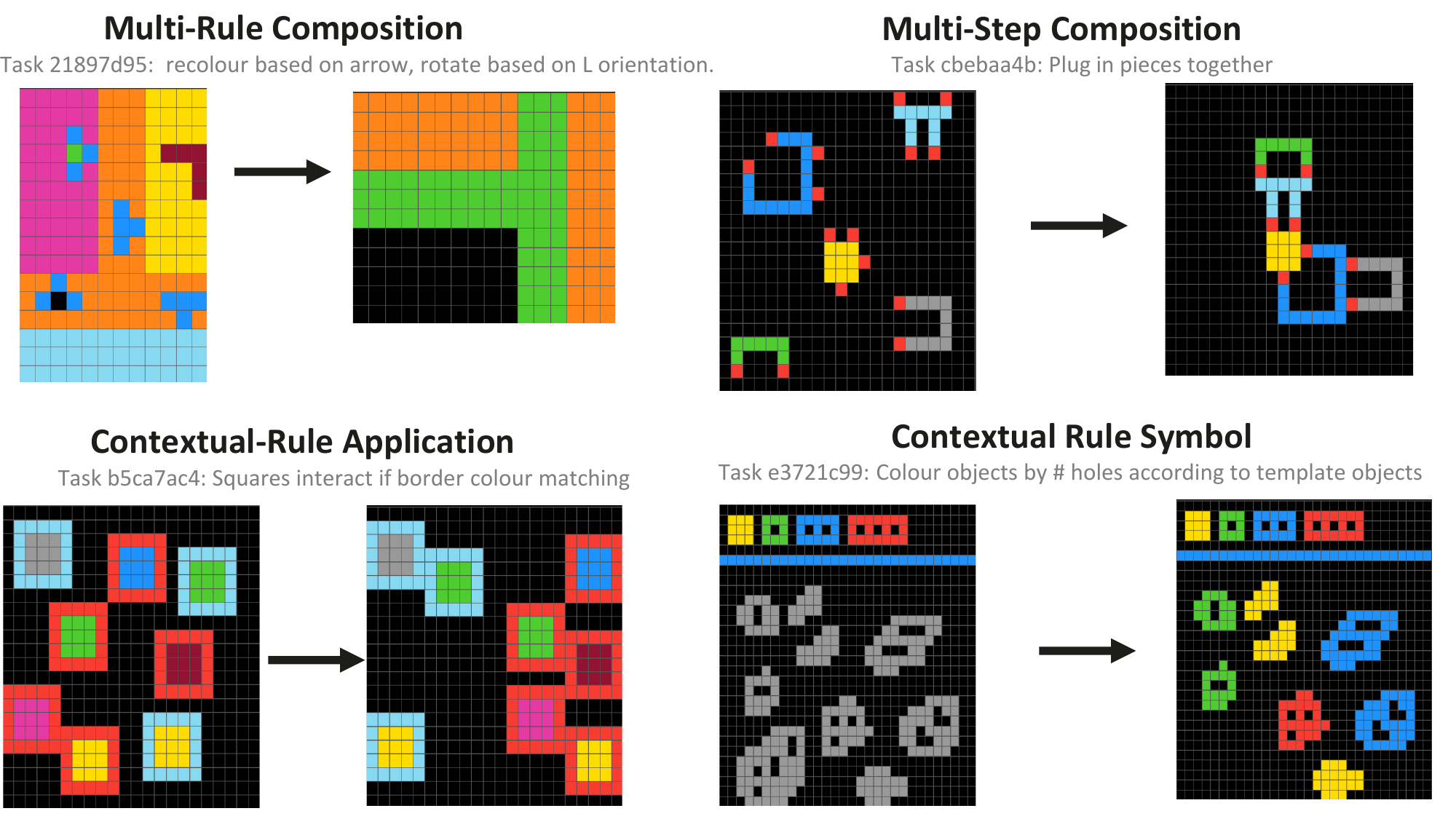}
    \caption{Various examples of the flavours of reasoning needed to solved ARC-AGI-2 tasks. }
    \label{fig:arc_task_types}
\end{figure}

Each task is defined by a singular, simplest latent algorithm that uniquely explains the transformation across the examples. Solving a task requires reverse-engineering this algorithm to determine the exact output grid for an unseen test input. Since each task is unique and shares only abstract ``core knowledge priors" (e.g., objectness, symmetry), models cannot rely on retrieving pre-learned solutions. See Figure \ref{fig:arc_task_types} for diverse examples of ARC-AGI-2 tasks. Instead, they must perform explicit new learning to induce the transformation rule. ARC-AGI remains a formidable challenge for AI, serving as a primary leaderboard for progress toward general intelligence \citep{chollet2025arc}.

While some early work attempted to solve ARC-AGI using discrete search \citep{xu2023graphs}, neural methods now dominate the literature. These approaches can be broadly categorized into \textit{program synthesis} and \textit{direct grid prediction}. A program synthesis approach is currently state of the art on ARC-AGI-2. \citet{Poetiq2025} leveraged top-tier closed-source LLMs to generate code solutions mapping input to output grid and use iterative prompting strategies to verify and refine solutions. Code generation tends to be not nearly as advantageous with open-source models, though \citet{pourcel2025self} recently achieved competitive results by constructing a large dataset reinterpreting its self-generated incorrect programs as correct programs for new tasks.

Our work primarily compares to open-source \textit{direct grid prediction} methods, which treat ARC as a token or image generation problem. The standard recipe for these models involves three components: (1) pretraining on large-scale synthetic datasets, (2) test-time fine-tuning on augmentations of the few-shot examples, and (3) search-based decoding with selection mechanisms like majority voting \citep{cole2025don, akyurek2024surprising, franzen2025product}. The winner of the 2025 ARC Challenge \citep{chollet2025arc}, NVARC \citep{sorokin2025nvarc}, followed this formula using \citet{franzen2025product}'s approach, with the critical innovation being a clever mechanism to leverage LLMs to generate a massive, diverse synthetic dataset for pretraining. 

The second-place solution \citep{franzen2025architects} is particularly relevant as it adapts Llada \citep{nie2025llada}, a pre-trained diffusion language model. However, their approach differs fundamentally from ours. They rely on a standard, fixed-depth transformer architecture with 4 billion parameters and repeatedly re-masking and denoising at test time. In contrast, our method uses diffusion to train a recursive, parameter-efficient loop from scratch.

Beyond architecture, recursion models differ from these LLM-based approaches in how they process few-shot examples. LLM approaches rely on \textit{in-context learning}, where all input-output pairs are concatenated into a single prompt. In contrast, HRM, TRM, and URM all train a shared \textit{task embedding} with the few-shot examples. The model then conditions on this embedding to process the test input. Conceptually, the forward pass in an LLM must simultaneously infer the transformation and apply it; in these models, the forward pass can be viewed as ``executing'' the learned task program (the embedding) on the provided input.

\section{DRM Reasoning Examples}\label{app:drm_vis}

We now show a few examples of DRM execution on test examples from ARC-AGI-2 Eval. During inference, DRM takes 16 denoising steps. We show the sequential predictions plus the remaskings for each step according to our noise schedule. We show the input and ground truth label at top of the image and the internal confidence $\hat{P}$(correct) across steps in the top right. Each prediction grid is annotated with the internal confidence and each remasking grid is annotated with remasking ratio---the fraction of cells being remasked.

\begin{figure}
\begin{center}
\includegraphics[height=0.88\textheight,keepaspectratio]{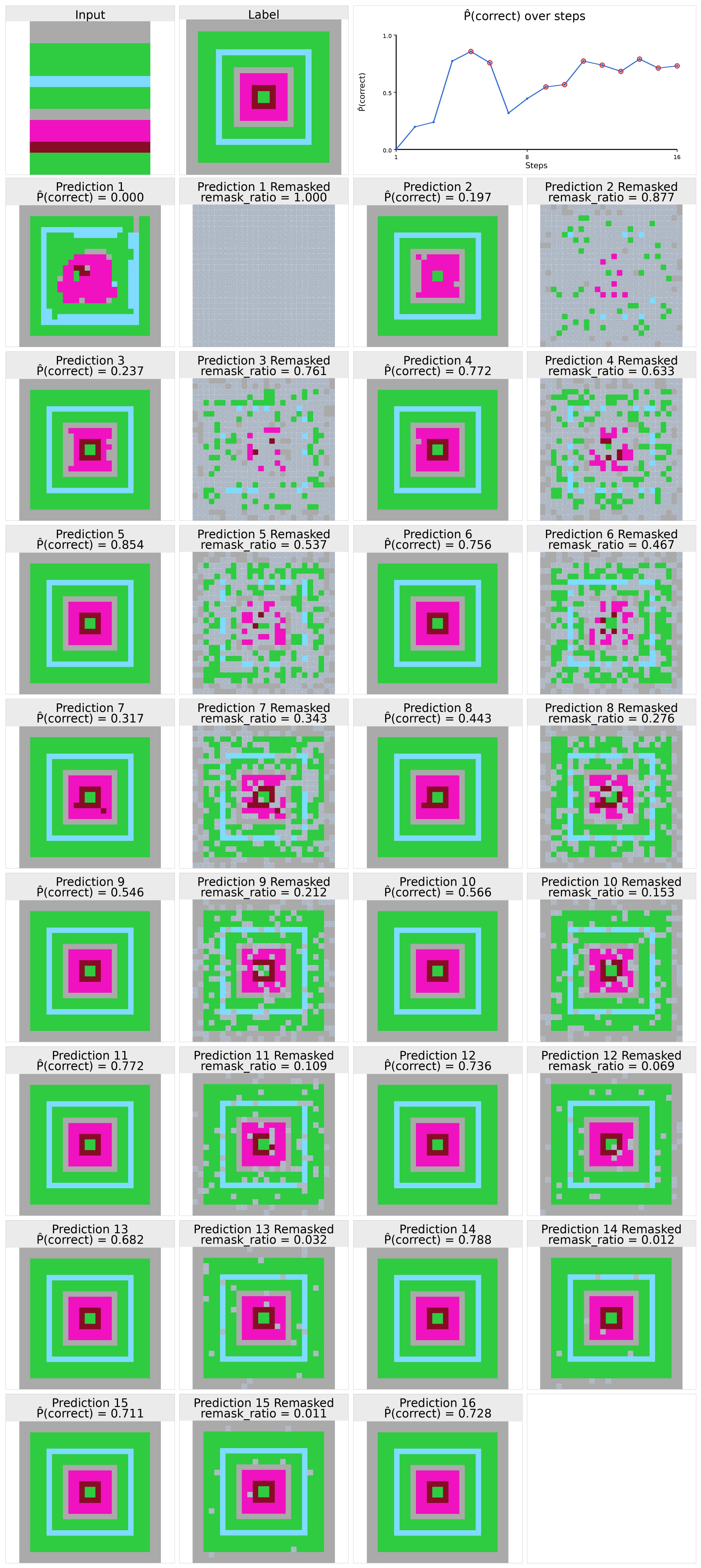}
\caption{Successful convergence on the ARC-AGI 2 \href{https://arcprize.org/play?task=45a5af55}{task \texttt{45a5af55}} of constructing nested squares with a certain colour order specified in the input. The model initially generates an incorrect square with a rough approximation of the correct answer and it refines it over several steps and eventually converges to the correct answer. Self-confidence $\hat{P}$(correct) is highly correlated with correctness indicating the model understood when it was correct and incorrect.}
\label{fig:45a5af55}
\end{center}
\end{figure}

\clearpage
\begin{center}
\includegraphics[height=0.88\textheight,keepaspectratio]{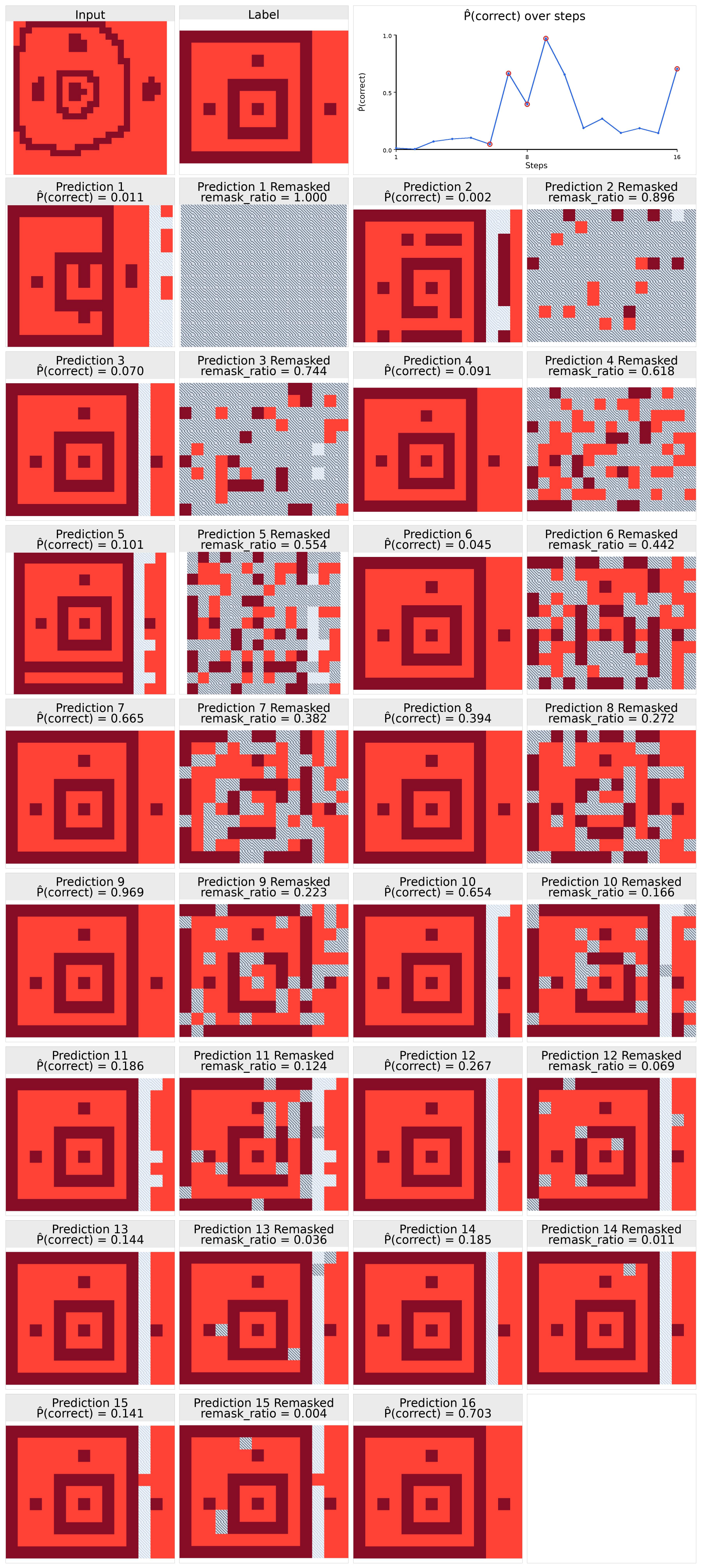}
\captionof{figure}{Successful solution on the ARC-AGI 2 \href{https://arcprize.org/play?task=2d0172a1}{task \texttt{2d0172a1}} of identifying core shapes from a rough ``sketch". Model is initially confused but eventually makes progress before oscillating between correct and incorrect solution and happens to settle on the correct answer in the final step.}
\label{fig:2d0172a1}
\end{center}

\clearpage
\begin{center}
\includegraphics[height=0.88\textheight,keepaspectratio]{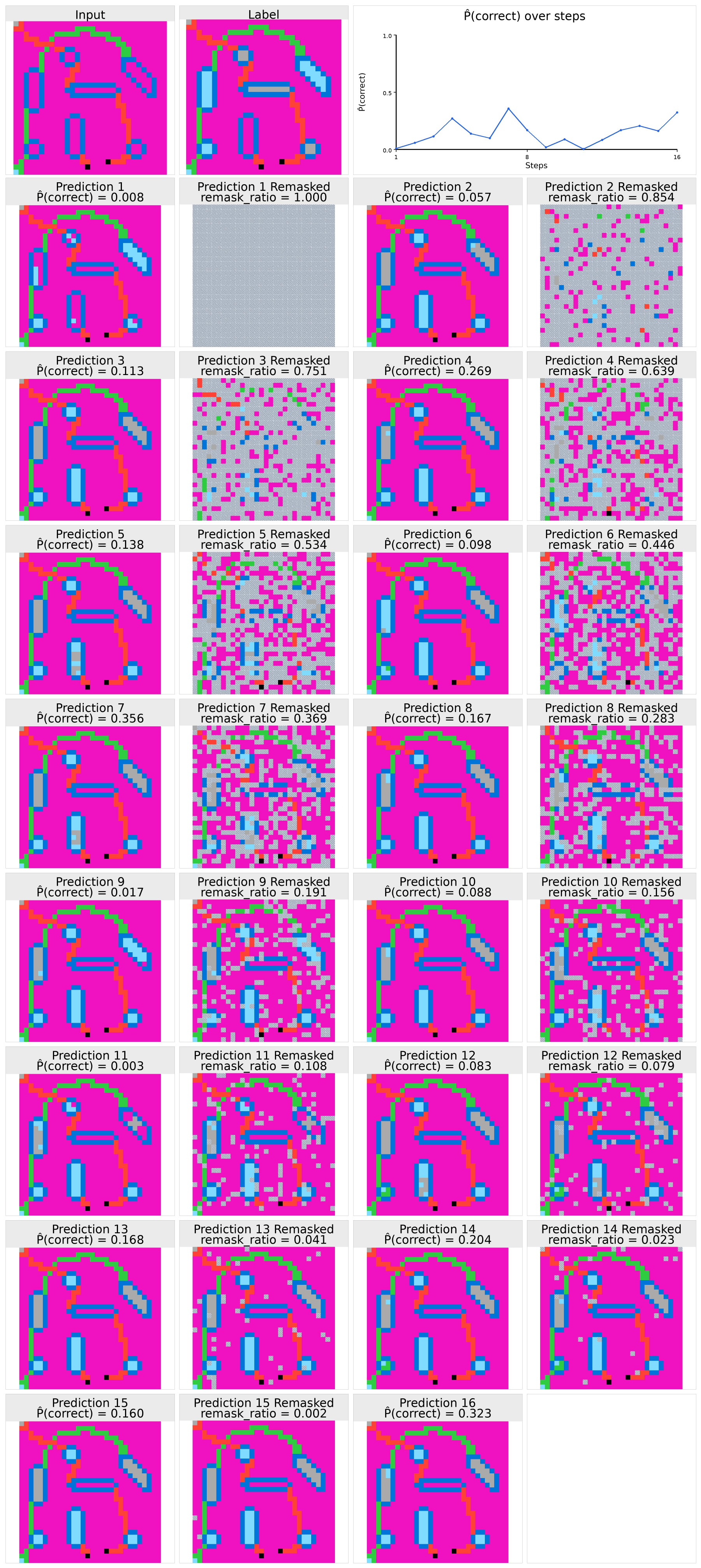}
\captionof{figure}{Failure to converge to a correct answer on a ARC-AGI 2 \href{https://arcprize.org/play?task=8b7bacbf}{task \texttt{8b7bacbf}} to fill in shapes corresponding to what colour path they lie on. The model seems to understand that the internals of the shapes must be filled but it clearly does not know what the colouring rule is. However, the model shows very low internal confidence indicating it knows it is confused.}
\label{fig:8b7bacbf}
\end{center}

\section{Pseudocode}
See Figure \ref{fig:pseudocode_comparison} for pseudocode comparing the training logic between TRM and DRM.

\begin{figure}[h!]
\centering
\footnotesize
\ttfamily
\setlength{\tabcolsep}{0pt}
\begin{tabular}{p{0.5\textwidth} p{0.5\textwidth}}
\toprule
\textrm{(a) TRM (Deep Supervision)} & \textrm{(b) DRM (Ours)} \\
\midrule

\begin{algorithmic}
\State \kw{def} trm\_rec(x, y, z, n=6, T=3):
\State \hspace{1em} \kw{with} torch.no\_grad():
\State \hspace{2em} \kw{for} j \kw{in} range(T-1):
\State \hspace{3em} \kw{for} i \kw{in} range(n):
\State \hspace{4em} z = net(x + y + z)
\State \hspace{3em} y = net(y + z)
\State \hspace{1em} \kw{for} i \kw{in} range(n):
\State \hspace{2em} z = net(x + y + z)
\State \hspace{1em} y = net(y + z)
\State \hspace{1em} \kw{return} (y, z), dec(y), Q(y)
\State
\State \kw{for} x, y\_true \kw{in} train\_data:
 \State \hspace{1em} \hlred{\# random initialization}
\State \hspace{1em} \hlred{y, z = y\_rand, z\_rand}
\State \hspace{1em} \hlred{\# iter over recursion windows}
\State \hspace{1em} \hlred{for step in N:}
\State \hspace{2em} x = input\_embed(x)
\State \hspace{2em} (y,z),y\_hat,q\_hat = trm\_rec(x,y,z)
\State \hspace{2em} loss = ce(y\_hat, y\_true)
\State \hspace{2em} loss += bce(q\_hat, (y\_hat==y\_true))

\State \hspace{2em} loss.backward()
\State \hspace{2em} opt.step()
\State \hspace{2em} \hlred{\# truncated backpropagration}
\State \hspace{2em} \hlred{opt.zero\_grad()}
\State \hspace{2em} \hlred{\# early-stopping}
\State \hspace{2em} \hlred{if q\_hat > 0: }
\State \hspace{3em} \hlred{break}
\end{algorithmic}

&

\begin{algorithmic}
\State \kw{def} drm\_rec(x, y, z, n=6, T=3):
\State \hspace{1em} \kw{with} torch.no\_grad():
\State \hspace{2em} \kw{for} j \kw{in} range(T-1):
\State \hspace{3em} \kw{for} i \kw{in} range(n):
\State \hspace{4em} z = net(x + y + z)
\State \hspace{3em} y = net(y + z)
\State \hspace{1em} \kw{for} i \kw{in} range(n):
\State \hspace{2em} z = net(x + y + z)
\State \hspace{1em} y = net(y + z)
\State \hspace{1em} \kw{return} (y, z), dec(y), Q(y)
\State
\State \kw{for} x, y\_true \kw{in} train\_data:
\State \hspace{1em} z = z\_rand
\State \hspace{1em} \hlgreen{\# noise target}
\State \hspace{1em} \hlgreen{tau = sample\_time\_step()}
\State \hspace{1em} \hlgreen{yn = add\_noise(y\_true, tau)}
\State \hspace{1em} yn = label\_embedding(yn)
\State \hspace{1em} x = input\_embedding(x)
\State \hspace{1em} (y,z),y\_hat,q\_hat = drm\_rec(x,yn,z)
\State \hspace{1em} loss = ce(y\_pred, y\_true)
\State \hspace{1em} loss += bce(q\_hat, (y\_hat==y\_true))
\State \hspace{1em} loss.backward()
\State \hspace{1em} opt.step()
\end{algorithmic}

\\
\bottomrule
\end{tabular}
\caption{Comparison of training logic between TRM and DRM (with TRM base). Text in red highlights logic that only appears in TRM, text in green highlights logic that only appears in DRM. (a) TRM uses truncated backpropagation. (b) Denoising Recursion uses a noise curriculum without truncated backpropagation.  }
\label{fig:pseudocode_comparison}
\end{figure}

\section{DRM/TRM relative strengths and weaknesses}

Overall, we found the tasks which are solved by each method to be highly correlated. We find that both are strongest on tasks where the transformation may be complex to uncover but easy to apply (e.g., fill in symmetry behind an occlusion: \href{https://arcprize.org/tasks/a6f40cea}{a6f40cea}) but both struggle on tasks with transformations requiring significant test-time search (e.g., plugging pieces together in the correct arrangement: \href{https://arcprize.org/tasks/cbebaa4b}{cbebaa4b}). To understand where each method's blind spots were relative to one another, we collected all tasks for each method that was solved at least once over the 1000 augmentations (pass@1000) on the largest pretraining scenario with 14M parameters.

Here are the 16 tasks that only DRM solved: \href{https://arcprize.org/tasks/0934a4d8}{0934a4d8}, \href{https://arcprize.org/tasks/2ba387bc}{2ba387bc}, \href{https://arcprize.org/tasks/2c181942}{2c181942}, \href{https://arcprize.org/tasks/2d0172a1}{2d0172a1}, \href{https://arcprize.org/tasks/4c3d4a41}{4c3d4a41}, \href{https://arcprize.org/tasks/4c416de3}{4c416de3}, \href{https://arcprize.org/tasks/4c7dc4dd}{4c7dc4dd}, \href{https://arcprize.org/tasks/64efde09}{64efde09}, \href{https://arcprize.org/tasks/8e5c0c38}{8e5c0c38}, \href{https://arcprize.org/tasks/a395ee82}{a395ee82}, \href{https://arcprize.org/tasks/a6f40cea}{a6f40cea}, \href{https://arcprize.org/tasks/b0039139}{b0039139}, \href{https://arcprize.org/tasks/b99e7126}{b99e7126}, \href{https://arcprize.org/tasks/b9e38dc0}{b9e38dc0}, \href{https://arcprize.org/tasks/c7f57c3e}{c7f57c3e}, \href{https://arcprize.org/tasks/dbff022c}{dbff022c}

We can imagine for many of these how a curriculum could help. For example, in task \href{https://arcprize.org/tasks/2c181942}{2c181942} (https://arcprize.org/tasks/2c181942), four different objects need to be placed in the correct positions. With masking, DRM will see many types of partial solutions of one, two, and three objects being correctly places and perhaps makes it more manageable to discover the correct transformation.

Here are the 8 tasks that only TRM solved: \href{https://arcprize.org/tasks/1ae2feb7}{1ae2feb7}, \href{https://arcprize.org/tasks/2b83f449}{2b83f449}, \href{https://arcprize.org/tasks/a251c730}{a251c730}, \href{https://arcprize.org/tasks/a47bf94d}{a47bf94d}, \href{https://arcprize.org/tasks/bf45cf4b}{bf45cf4b}, \href{https://arcprize.org/tasks/da515329}{da515329}, \href{https://arcprize.org/tasks/dd6b8c4b}{dd6b8c4b}, \href{https://arcprize.org/tasks/edb79dae}{edb79dae}

In Appendix \ref{app:maze_sudoku}, we show experiments where TRM outperformed DRM at Sudoku. The tasks where TRM is better make sense in light of this. Similar to Sudoku, 5/8 tasks require an unusual degree of test-time search i.e., even if you knew the transformation (rules of the task), it requires some thinking to apply. For example, task \href{https://arcprize.org/tasks/dd6b8c4b}{dd6b8c4b} (https://arcprize.org/tasks/dd6b8c4b) requires tracing each of the brown pixels to see if there is a viable path to the centre region. Because DRM is not trained over very long TBPTT sequences, it may not learn to build reasoning over very long reasoning traces as is required in tasks like this.

\section{Training Details}
See Table \ref{tab:training-hparams} for training details and Table \ref{tab:model-hparams} for archicture details.

\begin{table}[h]
\centering
\begin{tabular}{lcccc}
\toprule
\textbf{Hyperparameter} & \textbf{Train-from-scratch} & \multicolumn{2}{c}{\textbf{Pretrain}} & \textbf{Finetune} \\
\cmidrule(lr){3-4}
& & \textbf{reARC} & \textbf{NVARC} & \\
\midrule
Epochs & 100k & 100k & 2k & 300k \\
LR & 1e-4 & 1e-4 & 1e-4 & 1e-4 \\
Puzzle\_emb\_lr & 1e-2 & 1e-2 & 1e-2 & 1e-2 \\
lr\_warmup\_steps & 2k & 10k & 2k & 2k \\
lr\_anneal & no & no & no & no \\
Batch size & 768 & 768 & 768 & 768 \\
Weight decay & 0.1 & 0.1 & 0.1 & 0.1 \\
EMA & True & True & True & True \\
Loss & CE & CE & CE & CE\\
Optimizer & AdamW & AdamW & AdamW & AdamW\\
\bottomrule
\end{tabular}%
\caption{TRM training hyperparameters for each stage.}
\label{tab:training-hparams}
\end{table}

\begin{table}[h]
\centering
\begin{tabular}{lcccc}
\toprule
\textbf{Hyperparameter} & \multicolumn{2}{c}{\textbf{TRM/SPRM/DRM}} & \textbf{TRM (loop untie)} & \textbf{Transformer} \\
\cmidrule(lr){2-3}
& \textbf{7M} & \textbf{14M} & & \\
\midrule
Layers & 2 & 2 & 2 & 10 \\
Warm-up Windows ($H$ cycles) & 3 & 3 & 3 & 1 \\
Gradient Loops ($L$ cycles) & 4 & 6 & 4(untied) & 1 \\
Hidden size & 512 & 768 & 512 & 768 \\
Num heads & 8 & 12 & 8 & 12 \\
Expansion & 4 & 4 & 4 & 4 \\
Max halt steps & 16 & 16 & 16 & 16 \\

\bottomrule
\end{tabular}%

\caption{Model/config hyperparameters across methods.}
\label{tab:model-hparams}
\end{table}

\section{Extended Results}

\subsection{Pass@k Results}

\begin{figure*}[ht]
  \centering
    \includegraphics[width=\linewidth]{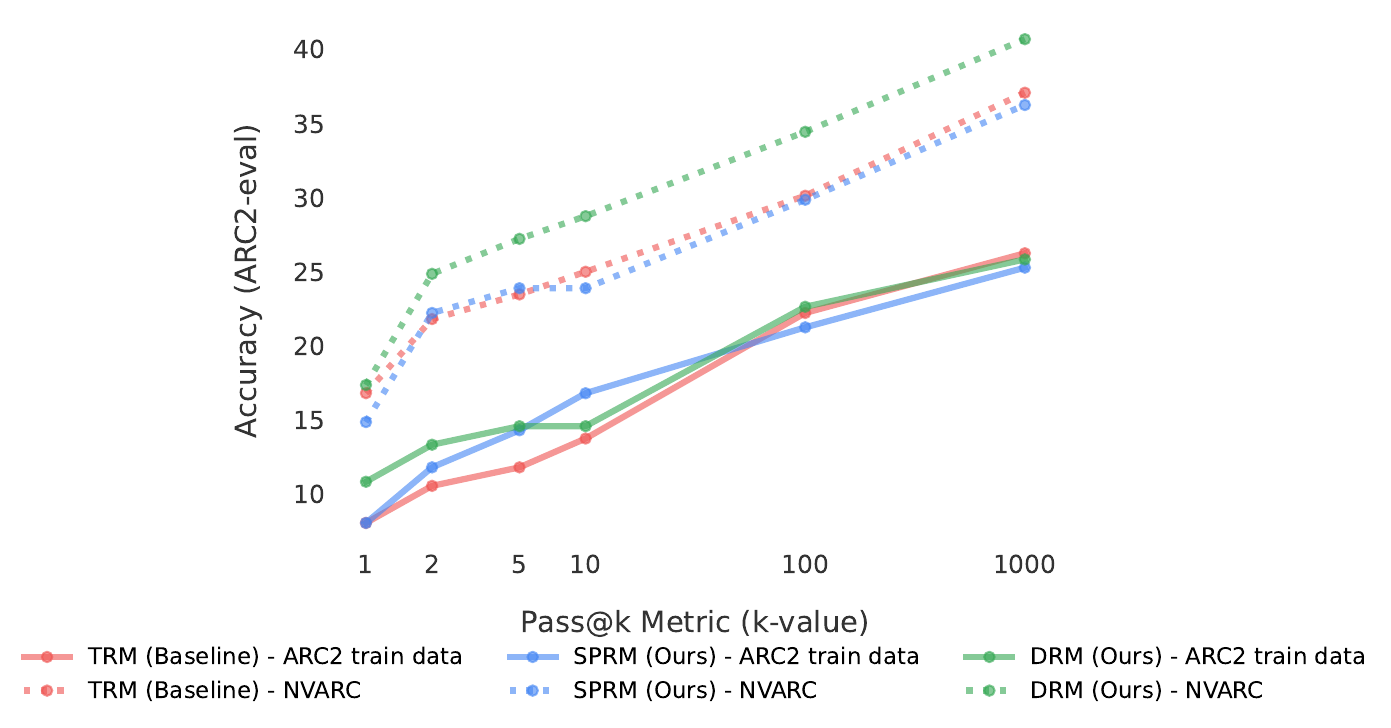}%
  \caption{Pass@k exact-match accuracy on ARC2-Eval for three methods (TRM baseline, SPRM, and DRM) with 14M parameters and pretrained with two scales of data (ARC2-train and NVARC)}
  \label{fig:pass@k}
\end{figure*}

Figure \ref{fig:pass@k} shows pass@k ARC2-Eval results for the 14M evaluations shown in Table \ref{tab:arc_results}. When pretrained on the largest NVARC dataset, DRM shows a clear improvement up to pass@1000, where it is $\sim5\%$ better than the TRM baseline. However, when pretraining on ARC2-train, TRM matches DRM performance at pass@100 and pass@1000. The gap between pass@1 and pass@1000 shows the potential for further improvement. We believe that generation of the correct output is much more difficult than identifying the correct output over a candidate set---i.e., we hypothesize humans would easily discriminate over correct and incorrect outputs.

\subsection{Signficance Results} \label{app:significance_results}

To test whether DRM and SPRM significantly improved over TRM, we performed a permutation test over the tasks under the null hypothesis of no performance difference. See Table \ref{tab:p_values} for p-values. The statistical power is much higher on ARC-Easy than ARC2-Eval given it contains 1000 as opposed to 120 tasks.

\begin{table}[htbp]
\centering
\begin{tabular}{llrrrr}
\toprule
 Test & \# params& \textbf{ARC-Easy} & \textbf{ARC2-eval} & \textbf{+reARC} & \textbf{+NVARC} \\
\midrule
SPRM$>$TRM & 7m  & $4.53 \times 10^{-6}$ & 0.0205 & 0.632 & 0.202 \\
DRM$>$TRM  & 7m  & $4.65 \times 10^{-5}$ & 0.0313 & 0.143 & 0.0772 \\
SPRM$>$TRM & 14m & 0.0449 & 0.266 & 0.744 & 0.500 \\
DRM$>$TRM  & 14m & 0.0418 & 0.124 & 0.251 & 0.183 \\

\bottomrule
\end{tabular}
\caption{P-values for permutation tests given the null hypothesis that TRM is better. }
\label{tab:p_values}
\end{table}

\subsection{Scaling with gradient loops and warm-up loops}

We investigated the effect of varying the number of recursive steps within a gradient window $k$ on the 7M ARC-Easy setting for both TRM and DRM. Following the TRM architecture, both methods perform two warm-up iterations with no gradient steps per loop prior to the $k$ gradient-bearing recursions. See Table \ref{tab:k_performance} for results.
\begin{table}[htbp]
    \centering
    \caption{Pass@2 exact match-accuracy on ARC-Easy across $k$ Values}
    \label{tab:k_performance}
    \begin{tabular}{lcccc}
        \toprule
        \textbf{Model} & \textbf{k=1} & \textbf{k=2} & \textbf{k=3} & \textbf{k=4} \\
        \midrule
        DRM & 31.8 & 37.8 & 45.1 & 50.5 \\
        TRM & 37.2 & 41.0 & 44.5 & 45.7 \\
        \bottomrule
    \end{tabular}
\end{table}

DRM suffeeds more at small $k$, with TRM outperforming DRM at $k=1,2$, the two methods reaching approximate parity at $k=3$, and DRM surpassing TRM at $k=4$. The marginal improvement for every additional recursive step is substantially larger for DRM than for TRM, suggesting that DRM benefits more from longer gradient windows. 

We note that standard masked diffusion corresponds to DRM with $k=1$ and no warm-up loops. The fact that DRM at $k=1$ achieves 31.8\% accuracy---compared to approximately 0\% for standard diffusion---indicates that the warm-up iterations, despite carrying no gradients, play a meaningful role in initializing the model's hidden state. For $k=1$, we show below the dependence of DRM and TRM on these warm-up steps.
\begin{table}[htbp]
    \centering
    \caption{Pass@2 exact match-accuracy on ARC-Easy across warm-up steps with $k$=1.}
    \label{tab:warmup_performance}
    \begin{tabular}{lccc}
        \toprule
        \textbf{Model} & \textbf{no warm-up} & \textbf{1 warm-up} & \textbf{2 warm-up} \\
        \midrule
        DRM & 0.0 (standard diffusion) & 17.7 & 31.8 \\
        TRM & 11.0 & 33.6 & 37.2 \\
        \bottomrule
    \end{tabular}
\end{table}
Both methods show very poor performance with no warmup loops but DRM is especially affected. We suspect that the warm-up loops allow better test-train alignment because the gradient loop starts with a self-generated intermediate state rather than starting directly from the masked target. Understanding the relative contributions of gradient loops and warm-up loops is an important direction of study.

\subsection{Noise Schedule and Mask Sampler Ablation} \label{app:noise_schedule_ablation}
We ablate the choice of noise schedule and mask sampler on the 7M ARC-Easy setting. We consider three noise schedules—linear, sigmoid, and cosine—paired with two mask sampling strategies: uniform random sampling and spatially correlated sampling using a 2D Gaussian kernel. The linear schedule weights all masking fractions equally, the sigmoid schedule heavily biases toward extreme (very high or very low) fractions, and the cosine schedule represents a middle ground with moderate bias toward both extremes.
Using uniform masking, the cosine schedule achieved 50.5\% accuracy, followed by linear at 46.2\% and sigmoid at 43.6\%. This is consistent with findings in prior diffusion work suggesting that a moderate bias toward high and low masking fractions is beneficial. We also compared uniform and spatially correlated masking under the cosine schedule; uniform achieved 50.5\% while spatially correlated achieved 49.0\%.

\subsection{TRM Extreme Robustness to Overfitting}

\begin{figure}[t]
  \centering
  \includegraphics[width=1\linewidth]{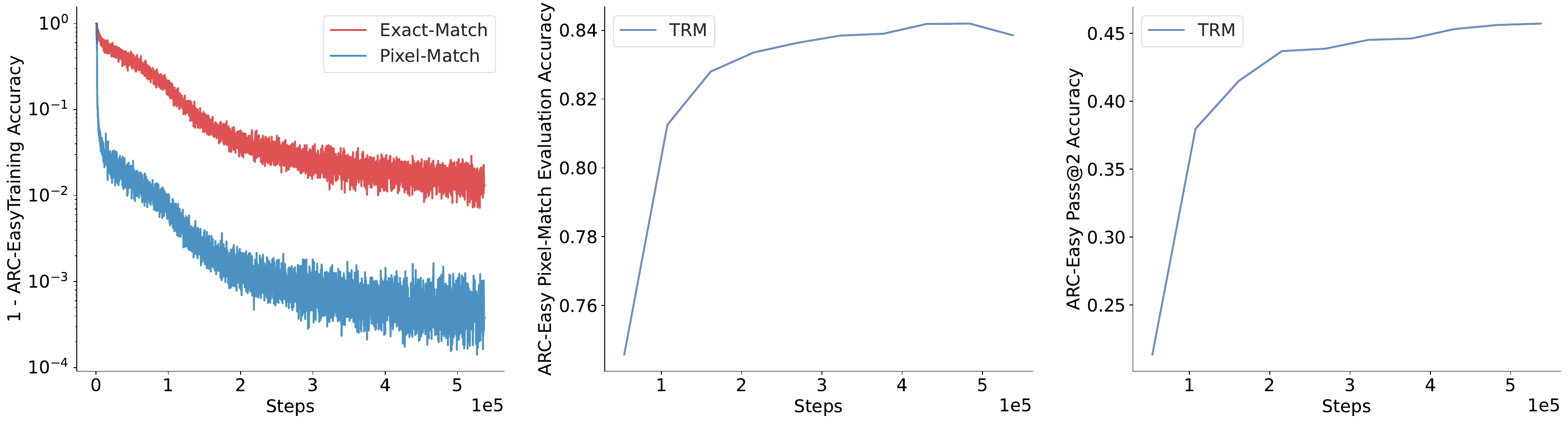}
  \caption{\textbf{Extreme robustness to overfitting.} TRM validation accuracy continues to improve even after training surpasses 99\% per-pixel accuracy across the grid. (left) 1--exact-match accuracy, 1--per-cell accuracy, (middle) validation per-cell accuracy, (right) validation pass@2. }
  \label{fig:nooverfit}
\end{figure}

Figure \ref{fig:nooverfit} illustrates an unusual training dynamic: validation continues to improve well after the model has effectively “solved” the training set. In particular, once training exceeds 99\% per-pixel exact accuracy across the entire output grid (100k steps), validation metrics keep rising from 38\% to 45\% between 100k and 500k steps. Note that here we are training on ARC2-train with $\sim$3k examples. Since our batch size is 768, four steps is virtually one epoch. This means that TRM does not overfit even after seeing a example about 100k times! Note that we do rotate through 1000 different ``augmentations" per data point (e.g., colour permutation, rotation), which does some of the work to avoid overfitting to basic artifacts. 

\begin{figure}[t]
  \centering
  \includegraphics[width=1\linewidth]{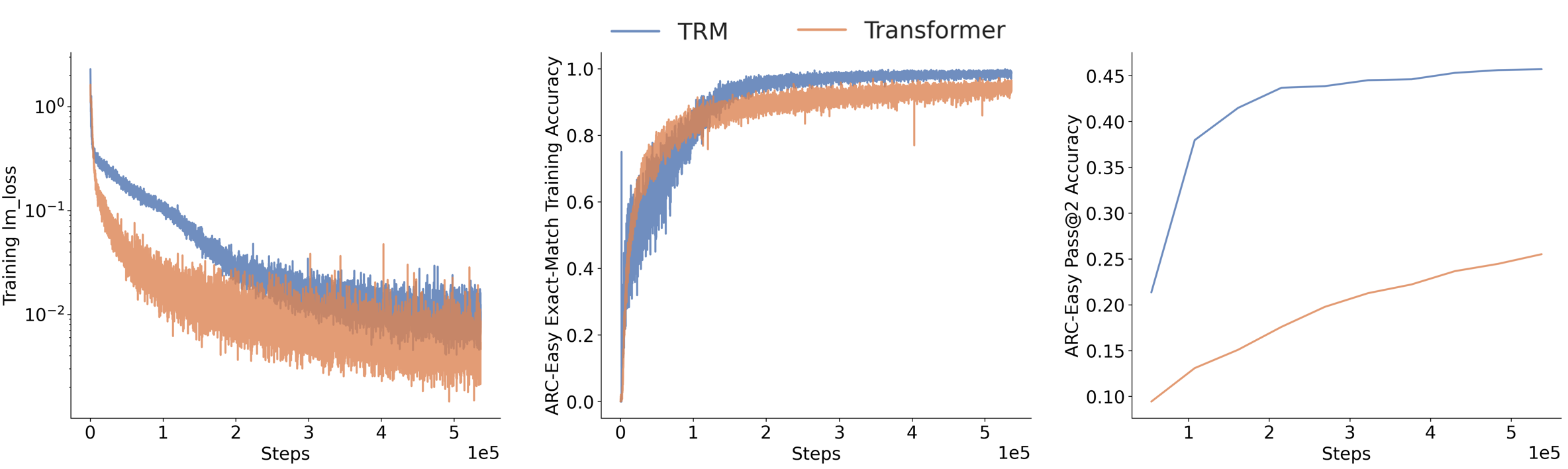}
  \caption{Unlooped transformer has better training loss than TRM baseline (left)) but worse training exact-match accuracy (centre) and higher evaluation pass@2 accuracy (right).}
  \label{fig:loop_v_unloop}
\end{figure}

When we instead stack layers without looping, the generalization is much worse. Figure \ref{fig:loop_v_unloop} shows that while training loss converges much faster (left) when we stack (Transformer), exact-match training performance of the looped model (TRM) quickly surpasses the stacked model (centre). This suggests that ``global" understanding is clearly worse when stacking. This translates to much better validation performance (right). 

\subsection{SPRM is a good regularization of input}
\label{subsec:sprm_regularization}

\begin{figure}[t]
    \centering
    \includegraphics[width=0.7\linewidth]{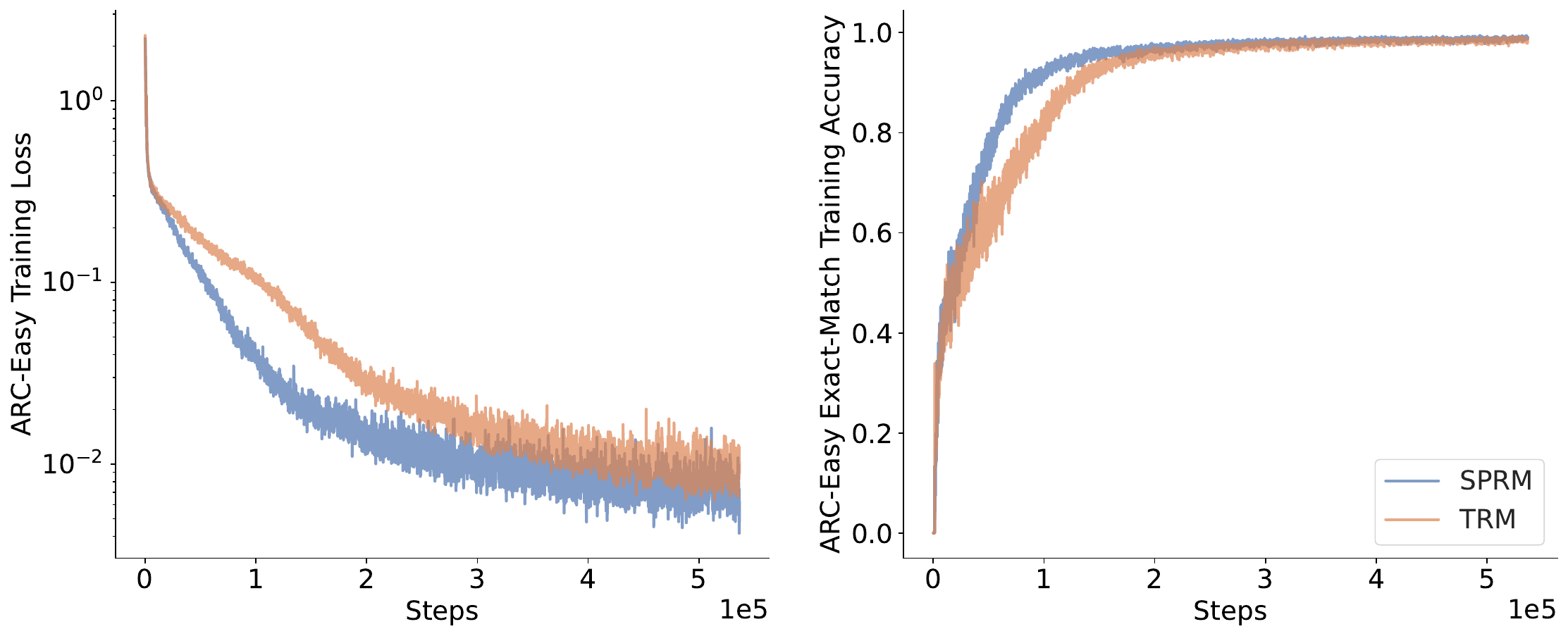}\hfill
    \caption{Training dynamics comparison between TRM and SPRM. (left) training exact accuracy and (right) training loss.}
    \label{fig:sprm_vs_trm_train}
\end{figure}

SPRM injects noise into the latent state that is carried over and used as the input to the next refinement step. We view noise injection as a regularizer, promoting generalization by increasing the diversity of training states because the same underlying example can generate many nearby latent inputs across refinement steps. 

Figure~\ref{fig:sprm_vs_trm_train} further indicates that SPRM outperform TRM from an optimization perspective. Adding noise actually improves convergence during training. By  ``thickening" the refinement trajectory, it may smooth the effective loss landscape since gradients are averaged over latent perturbations rather than tied to one precise hidden trajectory. It is counterintuitive how adding noise could improve training loss and accuracy but this does align with prior findings on RNNs that injecting noise into intermediate recurrent states can promote a contractive, stable recursion \citep{lim2021noisy}. We also note that noising intermediate steps is similar to Variational RNNs \citep{chung2015recurrent} which introduce latent random variables into the recurrence. Prior work on RNNs showed that noise injection can outperform Dropout \citep{dieng2018noisin}.

\subsection{Latent ``scratchpad'' is not needed}

\begin{figure}[t]
  \centering
  \includegraphics[width=1\linewidth]{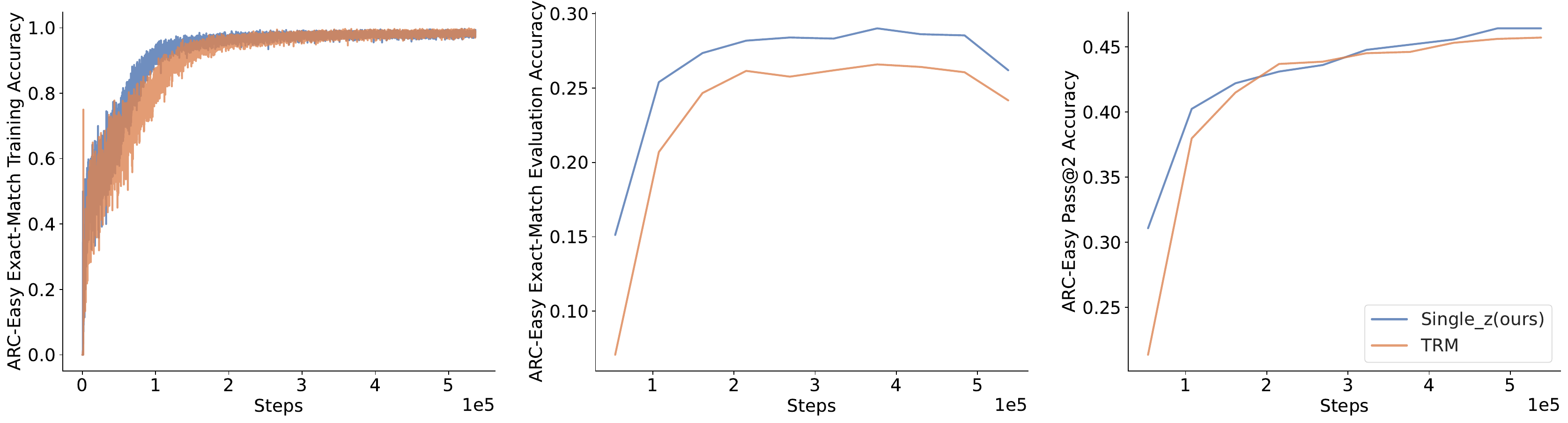}
  \caption{Single\_z (no scratchpad latent state) performed as good as the TRM baseline.}
  \label{fig:singlez}
\end{figure}

HRM \citep{wang2025hierarchical} maintains two latent states: a low-level state $z_L$ that repeatedly processes the input signal, and a high-level state $z_H$ that acts as a slower, summarizing carrier across cycles. $z_L$ is essentially a scratchpad state that continually is carried forward while $z_H$ is decoded into a solution. TRM adopts a similar two-latent-state design but simplifies the architecture by tying the high-level update to the same network used for low-level transitions. The HRM pattern:
\[
z_L \leftarrow f_L(z_L,\, z_H + x), \quad z_H \leftarrow f_H(z_H,\, z_L)
\]
becomes an update where both transitions reuse a single module $f$:
\[
z_L \leftarrow f(z_L,\, z_H + x), \quad z_H \leftarrow f(z_H,\, z_L).
\]

A natural question is whether this scratchpad $z_L$ is truly necessary.
In \citet{jolicoeur2025less}'s ablations, they test a more aggressive simplification that only depends on $z_H$
\[
z_H \leftarrow f(z_H, x), \qquad z_H \leftarrow f(z_H).
\]
which performed substantially worse. In contrast, our ablation shows that the scratchpad $z_L$ is not needed if we get rid of this ``summarization'' step:
\[
z_H^{(t+1)} \leftarrow f\!\left(z_H^{(t)},\, x\right).
\]
Pass@2 exact-match accuracy remains on par with the baseline as shown in Figure~\ref{fig:singlez} on ARC2-Easy. This suggests that the performance drop observed in their ablation is not due to removing $z_L$, but rather due to introducing an extra latent-only transition $z_H \leftarrow f_H(z_H)$.

This simplification was also made by the contemporaneous URM \citep{gao2025universal} and we argue this is cleaner since the recursion becomes perfectly homogeneous across iterations.

\subsection{Maze and Sudoku} \label{app:maze_sudoku}
To assess the generality of DRM beyond ARC-AGI, we evaluate on the Maze and Sudoku benchmarks studied by \citet{jolicoeur2025less}.

On the Maze task, DRM (where we mask maze path tokens) achieved 81\% accuracy compared to 78\% for TRM. We note that this evaluation likely underestimates the true performance of both methods, as we identified several instances in which TRM and DRM produced valid maze paths that differed from the ground-truth solution. Since the current evaluation protocol requires an exact match to the reference path, these correct but alternative solutions are scored as failures.

On Sudoku, DRM achieved 20\% accuracy compared to 87.9\% for TRM. We attribute this gap to the qualitative difference between Sudoku and typical ARC tasks. Sudoku puzzles are characterized by rules that are straightforward to understand but solutions that require extensive test-time search. Empirically, we observed that when TRM training on ARC converges, the average number of gradient windows required to solve training problems is just above one (1.2), whereas on Sudoku this number increases to approximately four. Because DRM does not train with truncated backpropagation through time (TBPTT) over multiple gradient windows, it may not learn to reason effectively over the very long sequential dependencies that Sudoku demands.

To further investigate this hypothesis, we conducted an additional experiment in which we corrupt the target as in DRM but then truncate and carry forward as in TRM. This hybrid approach achieved 85.7\% accuracy (compared to 87.9\% for standard TRM), suggesting that the carry-forward mechanism has importance for Sudoku that it does not have on ARC. This result indicates training DRM over longer gradient windows may help to close the performance gap.

We find corroborating evidence for this explanation in the ARC results. On ARC2-Eval, there are 8 tasks that TRM solves at pass@1000 but DRM never does. Five of these tasks---1ae2feb7, 2b83f449, a47bf94d, da515329, and dd6b8c4b---are Sudoku-like in that they require test-time search. Conversely, there are 16 tasks that DRM solves but TRM never does; many of these are complex, but none require a comparable degree of ``combinatorial" search.

When interpreting these results, it is important to consider that a 20\% vs 80\% gap on ARC corresponds to solving four times as many unique tasks, whereas on Sudoku it corresponds to solving four times as many instances of effectively the same task. We observed a sudden phase transition during TRM training on Sudoku, where accuracy jumped from approximately 20\% to 80\%. This behavior is consistent with the hypothesis that the Sudoku benchmark is best viewed as a single task of roughly uniform difficulty: once the model crosses a capability threshold, performance increases simultaneously across all instances. Unlike ARC, where many heterogeneous tasks produce a smoothly increasing performance curve, the uniform nature of Sudoku may cause the performance gap between DRM and TRM to appear larger than the underlying capability difference.

\end{document}